%% file: main.tex
\PassOptionsToPackage{table}{xcolor}
\documentclass{article}

\usepackage{microtype}
\usepackage{graphicx}
\usepackage{subcaption}
\usepackage{booktabs} 
\usepackage{tabularx} 
\usepackage{pifont} 
\usepackage{xurl} 
\usepackage{enumitem} 

\usepackage{hyperref}

\usepackage[preprint]{icml2026}

\usepackage{amsmath}
\usepackage{amssymb}
\usepackage{mathtools}
\usepackage{amsthm}

\usepackage[most]{tcolorbox}
\tcbuselibrary{listings}

\usepackage{tikz}
\usepackage{pgfplots}
\pgfplotsset{compat=1.18}
\usetikzlibrary{shapes,arrows,positioning,calc,fit,backgrounds,shadows,shadows.blur}
\pgfdeclarelayer{background}
\pgfsetlayers{background,main}

\definecolor{colorOurs}{RGB}{60, 100, 180}    
\definecolor{colorAccent}{RGB}{255, 111, 97}  
\definecolor{colorProp}{RGB}{47, 62, 70}      
\definecolor{colorBG}{RGB}{250, 250, 250}     
\definecolor{colorGrid}{RGB}{230, 230, 230}   
\definecolor{colorTextMuted}{RGB}{140, 140, 140} 
\definecolor{colorStroke}{RGB}{200, 200, 200} 

\newtcolorbox{promptbox}[1]{
    breakable,
    enhanced,
    before skip=12pt,
    after skip=12pt,
    colback=white,
    colframe=colorGrid,
    boxrule=0.5pt,
    sharp corners,
    fonttitle=\sffamily\bfseries\tiny,
    coltitle=colorOurs,
    title=\MakeUppercase{Prompt}: #1,
    attach title to upper,
    after title={\par\smallskip\hrule\smallskip},
    overlay={
        \fill[colorOurs] (frame.north west) rectangle ([xshift=2.5pt]frame.south west);
    },
    left=5mm,
    right=3mm,
    top=2mm,
    bottom=2mm
}

\definecolor{R1color}{rgb}{0.85, 0.1, 0.85} 
\definecolor{R2color}{rgb}{0.0, 0.7, 0.7}   
\definecolor{R3color}{rgb}{0.2, 0.2, 0.9}   

\usepackage{ulem}

\tikzset{
    avenir_font/.style={font=\sffamily\scriptsize, text=colorProp},
    avenir_bg/.style={fill=colorBG, draw=colorGrid, line width=0.4pt, rounded corners=6pt},
    card/.style={
        rectangle,
        rounded corners=4pt,
        align=center,
        inner sep=6pt,
        draw=colorStroke,
        line width=0.4pt,
        fill=white
    },
    hero_card/.style={
        card,
        draw=colorOurs!60,
        fill=colorOurs!5,
        font=\sffamily\bfseries\scriptsize,
        text=colorOurs
    },
    step_node/.style={
        rectangle,
        rounded corners=2pt,
        align=center,
        draw=colorGrid,
        line width=0.4pt,
        fill=white,
        font=\sffamily\scriptsize,
        inner sep=3pt
    },
    modern_arrow/.style={->, >=stealth, line width=0.5pt, draw=colorTextMuted},
    modern_arrow_bold/.style={modern_arrow, draw=colorOurs, line width=0.8pt},
    modern_arrow_dashed/.style={modern_arrow, dashed},
    tag/.style={
        font=\sffamily\tiny\bfseries,
        inner sep=2pt,
        rounded corners=2pt,
        fill=colorBG,
        draw=colorStroke,
        line width=0.4pt,
        text=colorTextMuted
    },
    our_tag/.style={
        tag,
        fill=colorOurs!15,
        draw=colorOurs!40,
        text=colorOurs
    },
    modern_grid/.style={draw=colorGrid, line width=0.3pt},
    modern_axis/.style={font=\sffamily\tiny, text=colorTextMuted},
    modern_label/.style={font=\sffamily\bfseries\scriptsize, anchor=south, yshift=2pt},
    modern_bar/.style={rounded corners=2pt, draw=none}
}

\usepackage[capitalize,noabbrev]{cleveref}

\theoremstyle{plain}

\theoremstyle{definition}

\theoremstyle{remark}

\usepackage[textsize=tiny]{todonotes}

\icmltitlerunning{Avenir-Web}

\begin{document}
\twocolumn[
  \icmltitle{Avenir-Web\texorpdfstring{\footnotemark[1]}{}: Human-Experience-Imitating Multimodal Web Agents with Mixture of Grounding Experts}

  \icmlsetsymbol{equal}{*}

  \begin{icmlauthorlist}
    \icmlauthor{Aiden Yiliu Li}{ucl,princeton}
    \icmlauthor{Xinyue Hao}{uoe}
    \icmlauthor{Shilong Liu}{princeton}
    \icmlauthor{Mengdi Wang}{princeton}
  \end{icmlauthorlist}

    \icmlaffiliation{ucl}{University College London, London, UK}
    \icmlaffiliation{uoe}{The University of Edinburgh, Edinburgh, UK}
    \icmlaffiliation{princeton}{Princeton University, Princeton, NJ, USA}
  \icmlcorrespondingauthor{Shilong Liu}{slongliu86@gmail.com}
  \icmlcorrespondingauthor{Mengdi Wang}{mengdiw@princeton.edu}
  \icmlkeywords{Multimodal Agent, Web Agent, Web Interaction, GUI Grounding, MLLM}

  \vskip 0.3in
]
\footnotetext[1]{'Avenir' is the French word for 'future', and also refers to the classic geometric sans-serif typeface known for its forward-looking aesthetic.}
\printAffiliationsAndNotice{ }  
\setcounter{footnote}{0}

\begin{abstract}

\input{sections/abstract}
\end{abstract}

\input{sections/introduction}
\input{sections/related_works}
\input{sections/methods}
\input{sections/experiments}
\input{sections/conclusion}

\bibliographystyle{icml2026}
\bibliography{references}

\input{sections/appendix}

\end{document}

%% file: sections/abstract.tex
Despite advances in Multimodal Large Language Models (MLLMs), autonomous web agents still struggle to reliably execute long-horizon tasks on complex and dynamic web interfaces.
Existing agents often suffer from inaccurate element grounding, the absence of site-specific procedural knowledge, and unstable long-term task tracking and memory, particularly when operating in complex Document Object Model (DOM) structures. To address these limitations, we introduce \textsc{Avenir-Web}, a web agent that achieves a new open-source state-of-the-art (SOTA) on the \textsc{Online-Mind2Web} benchmark in real-world deployment. \textsc{Avenir-Web} leverages Mixture of Grounding Experts (MoGE), Experience-Imitation Planning (EIP) for incorporating procedural priors, and a Task-Tracking Checklist combined with Adaptive Memory to enable robust and seamless interaction across diverse UI paradigms. 
We evaluate \textsc{Avenir-Web} on \textsc{Online-Mind2Web}~\cite{xue2025an}, a rigorous benchmark of live and user-centered web tasks. Our results demonstrate that \textsc{Avenir-Web} significantly surpasses prior open-source agents and attains performance parity with top-tier proprietary models, thereby establishing a new open-source state-of-the-art for reliable web agents on live websites.

%% file: sections/introduction.tex
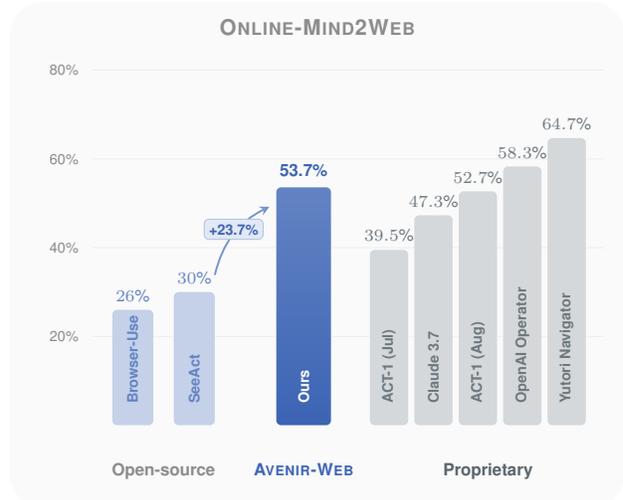
\begin{figure}[t]
    \centering
    \resizebox{\linewidth}{!}{
    \begin{tikzpicture}[avenir_font, scale=1.0, transform shape]
        \fill[colorBG, rounded corners=14pt] (-1.2, -1.2) rectangle (7.8, 6.2);
        
        \foreach \y/\l in {20/20\%, 40/40\%, 60/60\%, 80/80\%} {
            \draw[modern_grid, opacity=0.6] (0, \y*0.065) -- (7.5, \y*0.065);
            \node[left, modern_axis, xshift=-2pt] at (0, \y*0.065) {\l};
        }

        \node[modern_label, anchor=north, text=colorTextMuted, font=\sffamily\bfseries\small] at (3.3, 6.0) {\textsc{Online-Mind2Web}};
        
        \begin{pgfonlayer}{background}
            \fill[gray!5, rounded corners=8pt] (-0.1, -1.0) rectangle (2.2, 5.8);
            \fill[colorOurs!10, rounded corners=8pt, draw=colorOurs!20, line width=0.5pt] (2.4, -1.0) rectangle (3.8, 5.8);
            \fill[colorProp!5, rounded corners=8pt] (4.0, -1.0) rectangle (7.6, 5.8);
        \end{pgfonlayer}

        \node[modern_label, anchor=base, text=colorTextMuted] at (1.05, -0.8) {Open-source};
        \node[modern_label, anchor=base, text=colorOurs] at (3.1, -0.8) {\textsc{Avenir-Web}};
        \node[modern_label, anchor=base, text=colorProp!80] at (5.8, -0.8) {Proprietary};
    
        \foreach \x/\h/\label in {0.6/2.6/Browser-Use, 1.5/3.0/SeeAct} {
            \pgfmathsetmacro{\hval}{\h*0.65}
            \path[modern_bar, fill=colorOurs!30] (\x-0.3, 0) rectangle (\x+0.3, \hval);
            \node[avenir_font, font=\sffamily\bfseries\tiny, text=colorOurs!80, rotate=90, anchor=west] at (\x, 0.2) {\label};
            \pgfmathsetmacro{\val}{\h*10}
            \node[avenir_font, text=colorOurs!80, anchor=south] at (\x, \hval) {\pgfmathprintnumber[fixed, precision=1]{\val}\%};
        }

        \begin{pgfonlayer}{background}
            \fill[colorOurs!20, rounded corners=2pt] (2.75, -0.05) rectangle (3.45, 3.534);
        \end{pgfonlayer}
        \path[modern_bar, top color=colorOurs!80, bottom color=colorOurs] (2.7, 0) rectangle (3.5, 3.484);
        \node[avenir_font, font=\sffamily\bfseries\tiny, text=white, rotate=90, anchor=west] at (3.1, 0.2) {Ours};
        \node[avenir_font, font=\sffamily\bfseries\scriptsize, text=colorOurs, anchor=south, yshift=1pt] at (3.1, 3.484) {53.7\%};

        \foreach \x/\h/\label in {4.35/3.95/ACT-1 (Jul), 5.0/4.73/Claude 3.7, 5.65/5.27/ACT-1 (Aug), 6.3/5.83/OpenAI Operator, 6.95/6.47/Yutori Navigator} {
            \pgfmathsetmacro{\hval}{\h*0.65}
            \path[modern_bar, fill=colorProp!20] (\x-0.28, 0) rectangle (\x+0.28, \hval);
            \node[avenir_font, font=\sffamily\bfseries\tiny, text=colorProp!70, rotate=90, anchor=west] at (\x, 0.2) {\label};
            \pgfmathsetmacro{\val}{\h*10}
            \node[avenir_font, text=colorProp!70, anchor=south] at (\x, \hval) {\pgfmathprintnumber[fixed, precision=1]{\val}\%};
        }

        \draw[modern_arrow_bold, bend left=25, draw=colorOurs!70] (1.8, 2.2) to node[our_tag, pos=0.5, yshift=2pt] {+23.7\%} (2.6, 3.2);
        
    \end{tikzpicture}}
    \caption{Performance of \textsc{Avenir-Web} on the \textsc{Online-Mind2Web}~\cite{xue2025an} benchmark (300 live tasks). The figure compares the success rate of our agent with existing open-source baselines and proprietary models. \textsc{Avenir-Web} achieves a \textbf{53.7\%} success rate, which is shown alongside specialized agents such as \textsc{ACT-1}~\cite{enhans}, \textsc{Operator}~\cite{openai2025introducing_operator}, and \textsc{Navigator}~\cite{navigator}.}\label{fig:teaser_perf}
\end{figure}
\section{Introduction}
\textit{See it. Say it. Sorted!}
In modern computing environments characterized by complex and dynamic graphical user interfaces, a web agent is defined as an autonomous system~\cite{deng2023mind2web, zheng2024seeact} capable of perceiving web user interfaces, interpreting natural language instructions, and executing multi-step action sequences within a standard browser environment. These agents facilitate end-to-end assistance for intricate workflows, ranging from flight booking and form completion to enterprise data extraction and application configuration. Recent advancements in general-purpose multimodal large language models (MLLMs), such as GPT-4o~\cite{openai_computer_using_agent}, Gemini~\cite{google_deepmind_gemini3pro}, and Claude~\cite{anthropic2025sonnet45}, have established formidable reasoning and agentic planning capabilities~\cite{yao2023reactsynergizingreasoningacting, shinn2023reflexion}, enabling sophisticated interpretation of web content via structured HTML or accessibility tree analysis. Concurrently, the emergence of specialized GUI models~\cite{hong2024cogagentvisuallanguagemodel, lu2024omniparserpurevisionbased, lin2024showuivisionlanguageactionmodelgui} (foundational backbones explicitly trained for native OS-level interaction) has significantly enhanced visual grounding accuracy and native computer-interaction reliability. This trend has been further extended by recent open-source scaling efforts and vision-language-action paradigms~\cite{liu2025scalecuascalingopensourcecomputer, lin2024showuivisionlanguageactionmodelgui}. Together, these developments have catalyzed the emergence of autonomous systems~\cite{he2024webvoyagerbuildingendtoendweb, gur2024a} capable of grounding complex natural language commands into executable browser actions on live websites.

Despite this rapid progress, current web agents still suffer from critical reliability bottlenecks in real-world deployment.
Overall, these limitations can be summarized into three major bottlenecks: inaccurate \textit{element grounding}~\cite{yang2023setofmarkpromptingunleashesextraordinary, wu-etal-2025-dimo}, the lack of site-specific procedural knowledge~\cite{gur2024a, furuta2024multimodalwebnavigationinstructionfinetuned}, and unstable \textit{long-term task tracking and memory}~\cite{xue2025an}.
(1) First, regarding \textit{element grounding}, relying on a single modality or failing to deeply fuse multiple modalities is often insufficient for modern web applications~\cite{deng2023mind2web, gur2024a}. Agents can break on iframes, canvas elements, shadow DOMs, and other non-standard structures, or miss fine-grained semantic attributes and scrolling context. 
(2) Second, regarding site-specific procedural knowledge, existing systems generally lack the ability to learn from human experience encoded in external resources such as human-authored online guides. Without such guidance, agents are forced into trial-and-error exploration~\cite{xue2025an}, which significantly increases token consumption and the risk of task incompletion due to reaching step limits or encountering irreversible navigation errors.
(3) Third, regarding \textit{long-term task tracking and memory}, many agents remain largely reactive and short-horizon~\cite{shinn2023reflexion, yao2023reactsynergizingreasoningacting}. When tasks span multiple pages and state transitions, their internal state representation degrades, subgoal progress becomes ambiguous, and the agent exhibits navigational drift and compounding execution errors.

To address these limitations, we introduce \textsc{Avenir-Web}, a web agent that achieves a new open-source state-of-the-art (SOTA) on \textsc{Online-Mind2Web}, targeting the three core bottlenecks in real-world deployment.
Specifically, \textsc{Avenir-Web} integrates \textit{Mixture of Grounding Experts (MoGE)} for robust element grounding, \textit{Experience-Imitation Planning (EIP)} for incorporating \text{external procedural priors}, and a combination of \textit{Task-Tracking Checklist} and \textit{Adaptive Memory} for \text{long-term task tracking and memory} (see Table~\ref{tab:related-work} for a comparison with existing agents).

(1) To improve element grounding, we develop \textit{Mixture of Grounding Experts (MoGE)} for web tasks (Figure~\ref{fig:comparison}), which prioritizes a visual-first grounding path using an MLLM. This approach emulates human behavior by interacting with the interface as a unified visual canvas, effectively resolving elements within complex structures like nested iframes that typically paralyze DOM-centric agents~\cite{zheng2024seeact, he2024webvoyagerbuildingendtoendweb}. For specific edge cases requiring high-precision manipulation, MoGE utilizes a semantic structural reasoner as a robust fallback.
Compared with DOM-centric pipelines that often require separate inferences for Action Generation and Action Grounding~\cite{zheng2024seeact}, MoGE typically emits an executable action in a single inference step, while triggering extra rounds only for delicate operations (e.g., selection) or when the initially grounded point is non-responsive.
(2) To incorporate site-specific knowledge, we propose \textit{Experience-Imitation Planning (EIP)} (Figure~\ref{fig:eip_effect}), which retrieves and comprehends human-authored online guides~\cite{gur2024a} to produce high-level, site-specific plans, avoiding expensive open-ended exploration and reducing reliance on transient parametric memory.
(3) To stabilize long-horizon execution, we further introduce a \textit{Task-Tracking Checklist} that explicitly records subgoal completion~\cite{deng2023mind2web} and preserves a structured task state across page transitions, mitigating navigational drift and memory decay.
Finally, we implement \textit{Adaptive Memory} (Figure~\ref{fig:memory_comparison}) to manage interaction history through \textit{Chunked Recursive Summarization} and \textit{Failure Reflection}~\cite{shinn2023reflexion}. This module distills long-horizon execution traces and immediate failures into a persistent summary buffer, ensuring the agent maintains strategic situational awareness and can reason over past errors without exceeding context constraints.

We rigorously evaluate \textsc{Avenir-Web} on \textsc{Online-Mind2Web}~\cite{xue2025an}, a challenging benchmark reflecting diverse, live web domains. Our experiments demonstrate that \textsc{Avenir-Web} achieves a \textbf{23.7\%} absolute improvement in task success rate over existing open-source baselines (Figure~\ref{fig:teaser_perf}), establishing a new open-source state-of-the-art for reliable web agents on live websites. To further promote accessibility, we also introduce a \textit{fully open-source} configuration utilizing the lightweight \textit{Qwen-3-VL-8B}. This version achieves a \textbf{25.7\%} success rate, rivaling the performance of earlier baselines that rely on significantly larger proprietary backbones such as GPT-4o. This result underscores the framework's capacity to empower compact open-source models with robust, industry-standard agentic capabilities. These results validate that unifying Experience-Imitation Planning, Mixture of Grounding Experts (MoGE), the Task-Tracking Checklist, and Adaptive Memory significantly enhances an agent's ability to navigate the complexity of the modern web.

In summary, our primary contributions are:
\begin{itemize}[leftmargin=*, nosep, itemsep=4pt, topsep=4pt]
    \item We introduce \textsc{Avenir-Web}, a new \textsc{Online-Mind2Web}'s opensource state-of-the-art web agent designed for resilient long-horizon execution on live websites.
    \item We proposed \textit{Mixture of Grounding Experts (MoGE)} for robust element grounding, \textit{Experience-Imitation Planning (EIP)} for incorporating external procedural priors, and a combination of \textit{Task-Tracking Checklist} and \textit{Adaptive Memory} for resilient long-term task tracking and memory.
    \item We demonstrate \textsc{Online-Mind2Web}'s opensource state-of-the-art performance on the \textsc{Online-Mind2Web}~\cite{xue2025an} benchmark, achieving a \textbf{53.7\%} success rate and substantially outperforming existing open-source baselines (Figure~\ref{fig:teaser_perf}). Furthermore, our framework allows a \textit{fully open-source} setup with a lightweight 8B model to reach the performance of earlier state-of-the-art baselines. We release the codebase to facilitate reproducible research.\footnote{Code is available at \url{https://github.com/Princeton-AI2-Lab/Avenir-Web}.}
\end{itemize}

%% file: sections/related_works.tex
\section{Related Works}

\begin{table*}[t]
\centering
\caption{Comparison of \textsc{Avenir-Web} with existing autonomous web agents across key functional dimensions. The table lists features such as Experience-Imitation Planning (EIP), Task-Tracking Checklist, and Adaptive Memory. We leave some entries blank when implementation details are unavailable.}\label{tab:related-work}
\scalebox{0.85}{
\setlength{\tabcolsep}{6pt}
\renewcommand{\arraystretch}{1.0}
\begin{tabular}{lccccc}
\toprule
\textbf{Agent}& \textbf{Open Source} & \textbf{Grounding} & \textbf{External Knowledge} & \textbf{Checklist} & \textbf{Adapt. Mem.} \\
\midrule
Navigator~\cite{navigator} & \ding{55} & Vision Only & --- & --- & --- \\
ACT-1~\cite{enhans} & \ding{55} & --- & --- & --- & --- \\
SeeAct~\cite{zheng2024seeact} & \ding{51} & DOM Only & \ding{55}& \ding{55}& \ding{55} \\
Agent-E~\cite{abuelsaad2024agente} & \ding{51} & DOM Only & \ding{55}& \ding{55}& \ding{55} \\
Browser-Use~\cite{browseruse} & \ding{51} & DOM Only & \ding{55}& \ding{55}& \ding{55} \\
\midrule
\rowcolor{colorOurs!15} Avenir-Web (Ours) & \textbf{\ding{51}} &  \textbf{Mixture} & \textbf{\ding{51}} & \textbf{\ding{51}} & \textbf{\ding{51}} \\
\bottomrule
\end{tabular}
}
\end{table*}

\paragraph{Autonomous Web Agents.}
The development of autonomous web agents has evolved from heuristic-based scripts to sophisticated multimodal systems. Early efforts primarily focused on leveraging HTML document structures or visual information for web interaction~\cite{deng2023mind2web,gur2024a,shaw2023pixels,furuta2024multimodalwebnavigationinstructionfinetuned,lee2023pix2struct,yan2023gpt4vwonderlandlargemultimodal}. Building on these foundations, several prominent agent systems have emerged, including \textsc{Navigator}~\cite{navigator}, \textsc{ACT-1}~\cite{enhans}, \textsc{SeeAct}~\cite{zheng2024seeact}, \textsc{WebVoyager}~\cite{he2024webvoyagerbuildingendtoendweb}, \textsc{Browser-Use}~\cite{browseruse}, \textsc{Operator}~\cite{openai2025introducing_operator}, \textsc{ColorBrowserAgent}~\cite{zhou2026colorbrowseragentintelligentguiagent}, \textsc{ScaleCUA}~\cite{liu2025scalecuascalingopensourcecomputer}, \textsc{IBM CUGA}~\cite{marreed2025enterprisereadycomputerusinggeneralist}, \textsc{WebOperator}~\cite{dihan2025weboperatoractionawaretreesearch}, \textsc{AgentSymbiotic}~\cite{zhang2025symbioticcooperationwebagents}, \textsc{Learn-by-Interact}~\cite{su2025learnbyinteractdatacentricframeworkselfadaptive}, and \textsc{WebPilot}~\cite{zhang2024webpilotversatileautonomousmultiagent}. These systems explore different trade-offs between open-source availability, grounding modalities, and planning strategies (see Table~\ref{tab:related-work}). While these works show promise, generalizing to diverse web environments remains challenging. Recent agent frameworks have explored the potential of powerful LMMs (e.g., GPT-4V, Gemini) as generalist agents, demonstrating zero-shot capabilities on live websites. \textsc{Avenir-Web} builds on these foundations by integrating an \textit{Experience-Imitation Planner} and \textit{Adaptive Memory} to handle the reliability bottlenecks of long-horizon tasks.

\paragraph{MLLMs and GUI Grounding.}
Advanced Multimodal Large Language Models (MLLMs) such as \textsc{Qwen3}~\cite{bai2025qwen3vltechnicalreport,xu2025qwen3omnitechnicalreport}, \textsc{Claude 4.5}~\cite{anthropic2025sonnet45}, \textsc{Seed-1.8}~\cite{bytedance2025seed1_8}, and \textsc{Gemini 3 Pro}~\cite{google_deepmind_gemini3pro} have transformed web agents into strategic planners capable of sophisticated task decomposition and self-correction via reasoning traces~\cite{yao2023reactsynergizingreasoningacting,shinn2023reflexion}. Modern architectures move beyond linear prompting to externalize latent reasoning through structured state monitors and reflection loops, mitigating error propagation in long-horizon navigation. Simultaneously, specialized GUI models, including \textsc{UI-TARS}~\cite{qin2025uitarspioneeringautomatedgui,wang2025uitars2technicalreportadvancing}, \textsc{CogAgent}~\cite{hong2024cogagentvisuallanguagemodel}, \textsc{OmniParser}~\cite{lu2024omniparserpurevisionbased}, \textsc{ShowUI}~\cite{lin2024showuivisionlanguageactionmodelgui}, \textsc{OpenCUA}~\cite{wang2025opencuaopenfoundationscomputeruse}, \textsc{MAI-UI}~\cite{zhou2025maiuitechnicalreportrealworld}, \textsc{EvoCUA}~\cite{xue2026evocuaevolvingcomputeruse}, \textsc{GTA1}~\cite{yang2025gta1guitesttimescaling}, and others~\cite{openai_computer_using_agent,hcompany2025holo2,lei2025textscguispotlightadaptiveiterativefocus,tianxiaction,gou2025uground}, serve as powerful backbones for precise coordinate-based grounding. Despite these advances, GUI agents still face challenges with context loss and detection errors, prompting iterative strategies like \textit{Chain-of-Ground}~\cite{li2025chainofgroundimprovingguigrounding,wu-etal-2025-dimo,hsieh2025zonui3blightweightvisionlanguagemodel} to enhance interaction robustness. Further research continues to address model hallucinations~\cite{Huang_2025} and refine reasoning-eliciting techniques such as Set-of-Mark (SoM) and Chain-of-Thought~\cite{yang2023setofmarkpromptingunleashesextraordinary,wei2023chainofthoughtpromptingelicitsreasoning}.

%% file: sections/methods.tex
\section{Methods}\label{headings}
\begin{figure*}[t]
\centering
\resizebox{0.95\textwidth}{!}{
\begin{tikzpicture}[
    node distance=0.5cm and 0.5cm,
    font=\sffamily\scriptsize,
    >=latex,
    box/.style={
        rectangle, 
        draw=black!60, 
        rounded corners=3pt, 
        align=center, 
        minimum height=2.6em, 
        inner sep=4pt, 
        fill=white, 
        thick
    },
    module/.style={
        box, 
        fill=blue!5, 
        draw=blue!40
    },
    core/.style={
        box, 
        fill=orange!5, 
        draw=orange!40, 
        line width=1.0pt, 
        font=\sffamily\scriptsize, 
        minimum height=3.2em, 
        minimum width=3.5cm
    },
    env/.style={
        box, 
        fill=gray!10, 
        draw=gray!40, 
        minimum height=3em, 
        minimum width=14cm, 
        text width=10cm, 
        align=center
    },
    input/.style={
        box, 
        fill=gray!5, 
        draw=gray!40, 
        minimum width=10cm
    },
    group/.style={
        rectangle, 
        draw=gray!30, 
        dashed, 
        inner sep=8pt, 
        rounded corners=6pt, 
        fill=gray!2
    },
    arrow/.style={
        ->, 
        thick, 
        draw=gray!60, 
        rounded corners=3pt
    },
    dashed_arrow/.style={
        arrow, 
        dashed
    }
]

\node [input] (input) {User Instruction \& URL};

    \node [box, fill=yellow!10, draw=yellow!60, below=0.6cm of input, dashed, text width=2.4cm] (web_know) {Web Knowledge\\
        \tiny (Search \& Online Docs)};
    \node [module, left=0.6cm of web_know, text width=2.8cm] (planner) {Experience-Imitation\\
        Planner\\ \tiny (Live Knowledge Synthesis)};
    \node [module, right=0.6cm of web_know, text width=2.8cm] (checklist_gen) {Checklist\\
        Generator\\ \tiny (Subgoal Decomposition)};
    
    \begin{pgfonlayer}{background}
        \node [group, fit=(planner) (checklist_gen) (web_know), inner sep=12pt] (init_group) {};
    \end{pgfonlayer}
    \node [anchor=north west, font=\sffamily\scriptsize\bfseries, text=gray!80, inner sep=1pt, xshift=4pt, yshift=-2pt, fill=white] at (init_group.north west) {Initialization};

    \node [core, below=0.8cm of init_group] (agent) {Core Agent\\
        \tiny (Main Model)};

\node [module, left=0.8cm of agent, text width=3.2cm] (memory) {Adaptive Memory\\
    \tiny (Trace Distillation \& Reflection)};
\node [module, right=0.8cm of agent, text width=3.2cm] (checklist) {Task-Tracking Checklist\\
    \tiny (Progress Verification Monitor)};

\node [module, below=0.6cm of memory, xshift=0.5cm, text width=4.6cm] (perception) {Visual Perception\\
    \tiny (Viewport Encoding)};
\node [module, below=0.6cm of checklist, xshift=-0.5cm, text width=4.6cm] (moge) {Mixture of Grounding Experts (MoGE)\\
    \tiny (Multimodal Grounding)};

\node [env, below=1.8cm of agent] (env) {Web Environment\\ \tiny (Live Browser Context)};

\begin{pgfonlayer}{background}
    \node [group, fit=(agent) (memory) (checklist) (perception) (moge) (env), inner sep=12pt] (loop_group) {};
\end{pgfonlayer}
\node [anchor=north west, font=\sffamily\scriptsize\bfseries, text=gray!80, inner sep=1pt, xshift=8pt, yshift=-6pt, fill=white] at (loop_group.north west) {Execution Loop};

\draw [arrow] (input.south) -- (input.south |- init_group.north);
\draw [arrow] (planner.south) |- ($(agent.north) + (-1.2, 0.5)$) -- ($(agent.north) + (-1.2, 0)$) node[pos=0.7, left, font=\tiny, xshift=-4pt, fill=white, inner sep=1pt] {Strategy};
\draw [arrow] (agent.north |- init_group.south) -- node[fill=white, inner sep=2pt, font=\tiny, pos=0.5] {Instruction} (agent.north);
\draw [arrow] (checklist_gen.south) |- ($(agent.north) + (1.2, 0.5)$) -- ($(agent.north) + (1.2, 0)$) node[pos=0.7, right, font=\tiny, xshift=4pt, fill=white, inner sep=1pt] {Checklist};

\draw [dashed_arrow] (web_know.west) -- (planner.east);

\draw [arrow, <->] (agent.west) -- (memory.east);
\draw [arrow, <->] (agent.east) -- (checklist.west);

\draw [dashed_arrow] (env.north -| perception.center) -- node[midway, left, font=\tiny, fill=white, inner sep=1pt] {Raw State} (perception.south);

\draw [arrow] (perception.east) -| ($(agent.south) + (-1.0, 0)$) node[pos=0.9, left, font=\tiny, fill=white, inner sep=1pt, xshift=-2pt] {Perception};
\draw [arrow] ($(agent.south) + (1.0, 0)$) |- (moge.west) node[pos=0.1, right, font=\tiny, fill=white, inner sep=1pt, xshift=2pt] {Action Intent};

\draw [arrow] (moge.south) -- node[midway, right, font=\tiny, fill=white, inner sep=1pt] {Operation} (moge.south |- env.north);

\draw [dashed_arrow] ($(env.north -| memory.west) + (-0.6, 0)$) |- (memory.west) node[pos=0.2, above, rotate=90, font=\tiny, yshift=4pt, fill=white, inner sep=1pt] {Operation Outcome};
\draw [dashed_arrow] ($(env.north -| checklist.east) + (0.6, 0)$) |- (checklist.east) node[pos=0.2, below, rotate=90, font=\tiny, yshift=4pt, fill=white, inner sep=1pt] {State Observation};

\end{tikzpicture}
}\caption{System architecture of \textsc{Avenir-Web}, featuring a decoupled strategic planning and execution framework. The \textit{Initialization phase} utilizes the \textbf{Experience-Imitation Planner (EIP)} to transform external procedural knowledge into a verifiable \textbf{Task-Tracking Checklist}. During the iterative \textit{Execution Loop}, the agent maintains strategic consistency through \textbf{Adaptive Memory} while the \textbf{Mixture of Grounding Experts (MoGE)} ensures robust element interaction via hierarchical visual-semantic grounding. A closed-loop feedback mechanism propagates environmental state observations back to the checklist and memory modules to prevent navigational drift in long-horizon tasks.}\label{fig:architecture}
\end{figure*}
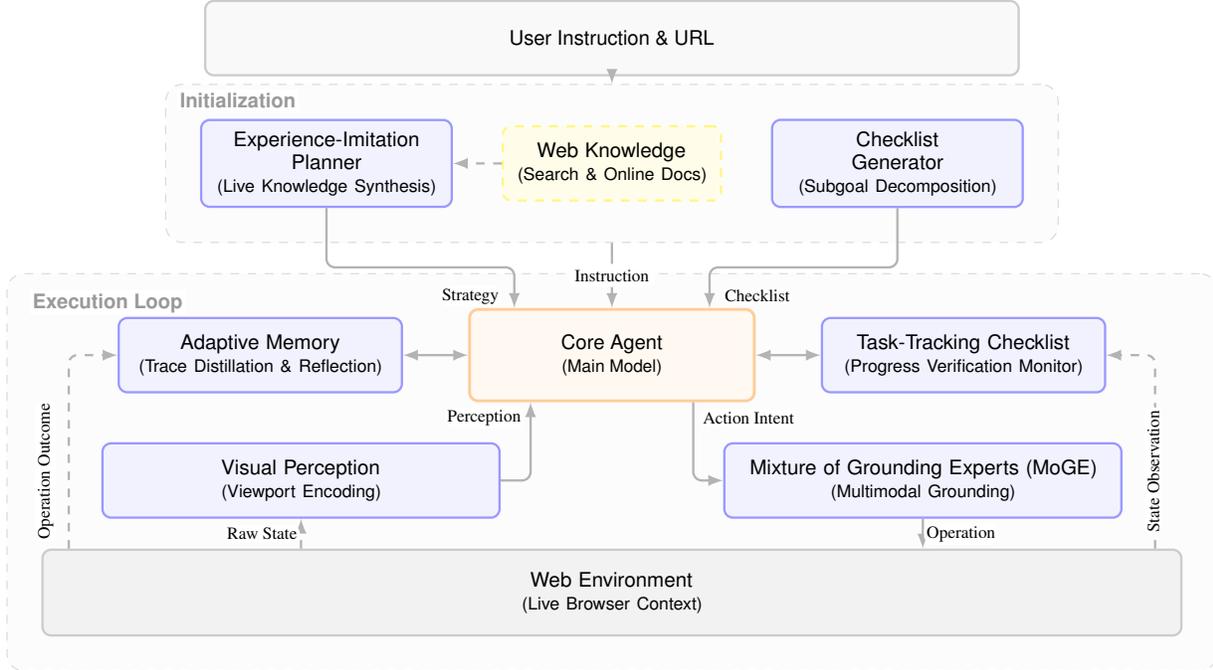
The \textsc{Avenir-Web} framework introduces a modular architecture designed to address the reliability bottlenecks of autonomous web navigation through four primary components: \textit{Experience-Imitation Planning (EIP)}, the \textit{Task-Tracking Checklist}, \textit{Adaptive Memory}, and the \textit{Mixture of Grounding Experts (MoGE)}. Our system distributes reasoning and execution responsibilities across specialized modules to ensure consistency, interpretability, and resilience in dynamic web environments. The overall design is structured around two functional phases that directly target the core bottlenecks: Initialization for strategic synthesis, and the Execution Loop for resilient task tracking and robust element interaction (Figure~\ref{fig:architecture}).

The system's execution pipeline follows a structured transition from strategic synthesis to iterative interaction. During the Initialization phase, the \textit{Experience-Imitation Planner} establishes a strategic roadmap by breaking down the user goal into a high-level plan and an initial \textit{Task-Tracking Checklist} of verifiable success criteria. Once initialized, the Action Agent enters a persistent Execution Loop to navigate the target interface. In each step, the agent integrates its current perception with the global strategy and historical context from \textit{Adaptive Memory} to determine the optimal next operation. This intent is then operationalized through the MoGE module, which prioritizes direct visual grounding using general-purpose multimodal models while leveraging semantic structural reasoning to resolve complex interface elements. Finally, the system updates the \textit{Task-Tracking Checklist} to monitor progress and detect execution anomalies, ensuring robust goal-oriented navigation.

\subsection{Experience-Imitation Planning}\label{sec:eip}

\begin{figure}[t]
\centering
\vspace{5pt}
\includegraphics[width=\columnwidth]{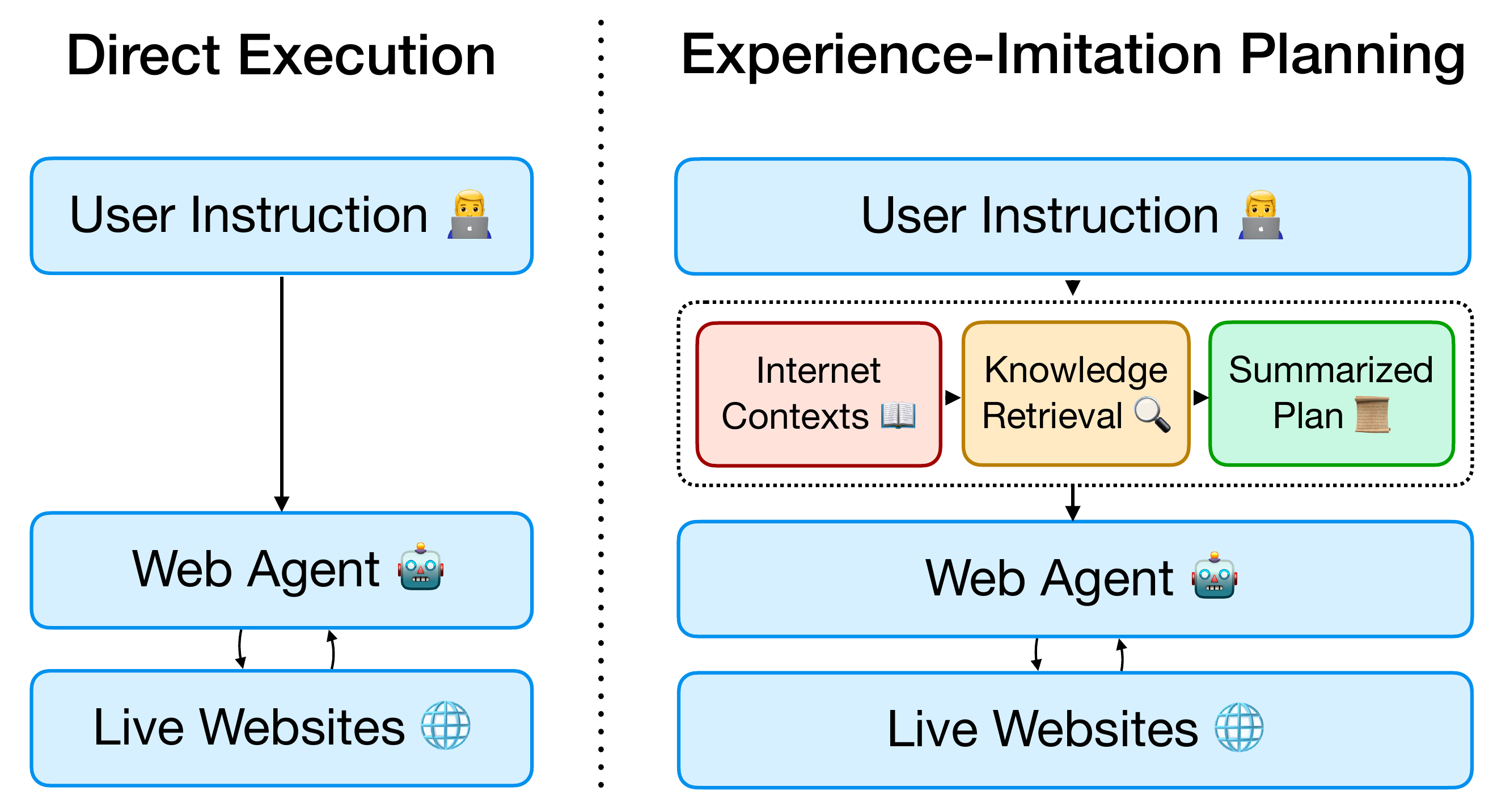}
\caption{Comparison of Experience-Imitation Planning (EIP). Without EIP, the agent executes instructions directly against the live website. With EIP, external how-to knowledge is searched and summarized into a site-specific plan that guides the agent's interaction.}
\label{fig:eip_effect}
\end{figure}

The Experience-Imitation Planner is a reasoning module that provides the system with procedural guidance by emulating human experience derived from online knowledge (Figure~\ref{fig:eip_effect}). Many web tasks require knowledge of site-specific workflows or familiarity with navigation patterns that are unique to the target website~\cite{gur2024a, deng2023mind2web}. To bridge this gap, the planner uses Claude 4.5 Sonnet~\cite{anthropic2025sonnet45} with its online search capability to retrieve site-specific resources such as forums, help centers, or user guides, identifying the most effective interaction sequences for a given goal. This strategic exploration phase is similar to paradigms employed by advanced multi-agent systems~\cite{zhang2024webpilotversatileautonomousmultiagent} and tree-search-based operators~\cite{dihan2025weboperatoractionawaretreesearch} that prioritize high-level roadmap synthesis over immediate reaction. 

Without such human-derived experience, agents are often forced into open-ended exploration, which carries significant disadvantages: it leads to excessive token consumption, risks reaching maximum step limits before task completion, and increases the likelihood of encountering irreversible navigation errors that can terminate the entire session. This procedural imitation allows the agent to anticipate common obstacles and prioritize high-success navigation paths, significantly reducing the exploration time required to solve unfamiliar tasks.

This module operates during the initialization phase to establish a strategic roadmap before the interaction loop begins. By accessing site-specific online resources, the planner synthesizes the user instruction and target URL into a high-level plan consisting of 2-4 imperative directives (see Figure~\ref{fig:petfinder_example} for an example). For instance, while a reactive agent might attempt to find a Careers link by searching the main navigation bar, the Experience-Imitation Planner identifies from the site's specific documentation that such links are located exclusively in the page footer, directing the agent to scroll to the bottom of the page immediately. This plan is restricted to abstract action descriptions rather than precise selectors to ensure it remains robust across different interface layouts. This approach allows the Action Agent to follow a proven procedural path, decoupling high-level strategy from low-level execution and preventing the system from becoming stuck in local navigation errors.

\input{sections/petfinder_example}

\subsection{Task-Tracking Checklist}
\begin{figure}[t]
\centering
\vspace{8pt}
\resizebox{\columnwidth}{!}{
\begin{tikzpicture}[
    node distance=0.2cm and 0.5cm,
    font=\sffamily\fontsize{11}{12}\selectfont,
    >=latex,
    box/.style={rectangle, draw=black!60, rounded corners=1pt, align=center, minimum height=1.3em, inner sep=3pt, fill=white, line width=0.5pt},
    module/.style={box, fill=blue!5, draw=blue!40},
    process/.style={box, fill=orange!5, draw=orange!40, text width=4.5cm},
    data/.style={box, fill=gray!5, draw=gray!40, text width=4.5cm},
    group/.style={rectangle, draw=gray!30, dashed, inner sep=5pt, rounded corners=4pt, fill=gray!2, line width=0.4pt},
    arrow/.style={->, line width=0.5pt, draw=gray!60, rounded corners=3pt}
]

\node [data] (task) {User Instruction};
\node [module, below=of task, text width=4.5cm] (gen_llm) {Outcome Generation};
\node [process, below=of gen_llm] (enforce) {Structural Refinement};
\node [data, below=of enforce] (init_checklist) {Initial Checklist ($C_0$)};

\draw [arrow] (task) -- (gen_llm);
\draw [arrow] (gen_llm) -- (enforce);
\draw [arrow] (enforce) -- (init_checklist);

\begin{pgfonlayer}{background}
    \node [group, fit=(task) (init_checklist), label={[anchor=south west, font=\sffamily\fontsize{12}{13}\selectfont]north west:1. Generation}] (gen_group) {};
\end{pgfonlayer}

\node [data, right=1.2cm of task, text width=7.0cm] (update_context) {Update Context ($a_t, o_t, S_t$)};
\node [module, below=of update_context, text width=7.0cm] (update_llm) {Checklist Synchronizer};
\node [data, below=of update_llm, text width=7.0cm] (updated_checklist) {Checklist $C_{t+1}$};
\node [box, below=0.3cm of updated_checklist, dashed, text width=7.0cm] (action) {Action Agent Execution ($a_{t+1}$)};

\draw [arrow] (update_context) -- (update_llm);
\draw [arrow] (update_llm) -- (updated_checklist);
\draw [arrow] (updated_checklist) -- (action);

\draw [arrow] (action.east) .. controls +(0.5,0) and +(0.5,0) .. (update_context.east);

\begin{pgfonlayer}{background}
    \node [group, fit=(update_context) (action), label={[anchor=south west, font=\sffamily\fontsize{12}{13}\selectfont]north west:2. Execution Loop}] (update_group) {};
\end{pgfonlayer}

\end{tikzpicture}
}
\vspace{3pt}
\caption{The Task-Tracking Checklist lifecycle. (1) Initialization generates atomic outcome states from the user instruction. (2) During the execution loop, a lightweight model iteratively updates the checklist based on action feedback and environment observations.}
\label{fig:checklist_flow}
\end{figure}
To maintain long-term goal focus and prevent repetitive failure loops, \textsc{Avenir-Web} utilizes a Task-Tracking Checklist that decomposes complex instructions into verifiable milestones~\cite{deng2023mind2web} (Figure~\ref{fig:checklist_flow}). This structured decomposition is critical for overcoming the compounding error problem observed in long-horizon web navigation benchmarks~\cite{xue2025an, he2024webvoyagerbuildingendtoendweb}. Figure~\ref{fig:petfinder_example} illustrates the parallel initialization of the Experience-Imitation Planning (EIP) and the Task-Tracking Checklist for a navigation task on \url{petfinder.com}. The checklist defines \textit{atomic outcome states} that remain valid regardless of the specific navigation path taken.

\textbf{Checklist Generation.} When a new task starts, the system establishes the initial success criteria. The Experience-Imitation Planner constructs a prompt that instructs the model to create 2--6 atomic items, each limited to 10 words and output in a structured JSON format. To ensure reliability, the raw output is passed through a refinement layer that splits complex sentences, deduplicates similar goals, and initializes all statuses to a pending state. If the model fails to provide a valid structure after retries, the system utilizes a hardcoded fallback consisting of general execution and completion checkpoints.

\textbf{Iterative Updating.} After every interaction step $t$, the checklist is synchronized with the environment. Let $C_t = \{(m_i, s_{i,t})\}_{i=1}^k$ be the checklist at step $t$, where $m_i$ is the $i$-th milestone and $s_{i,t} \in \{\text{P, IP, C, F}\}$ is its status (Pending, In Progress, Completed, Failed). The updated checklist $C_{t+1}$ is computed as:
\begin{equation}
    C_{t+1} = \mathcal{F}_{\theta}(C_t, a_t, o_t, S_t)
\end{equation}
where $a_t$ is the executed action, $o_t$ is the observed outcome, and $S_t$ is the current page state. This update is performed by a lightweight \textit{Qwen-3-VL-8B}~\cite{bai2025qwen3vltechnicalreport} model tasked with updating exactly one item most directly affected by the action. This selective update strategy minimizes latency and ensures that the checklist remains a reliable source of truth for the Action Agent throughout the execution loop.

\begin{figure}[t]
\centering
\resizebox{\columnwidth}{!}{
\begin{tikzpicture}[
    avenir_font,
    header/.style={font=\sffamily\fontsize{8.5}{9.5}\selectfont, align=center},
    card/.style={rectangle, draw=gray!20, fill=white, minimum width=2.9cm, minimum height=1.6cm, align=center, font=\sffamily\fontsize{7.5}{8.5}\selectfont},
    our_card/.style={card, draw=colorOurs!40, fill=colorOurs!2, line width=0.8pt},
    code_box/.style={card, minimum height=0.8cm, font=\sffamily\fontsize{7.5}{8.5}\selectfont, fill=gray!2},
    our_code_box/.style={code_box, draw=colorOurs!30, fill=colorOurs!8},
    flow/.style={-stealth, draw=gray!30, line width=0.5pt},
    our_flow/.style={flow, draw=colorOurs!60, line width=0.8pt}
]

    \fill[gray!2] (-1.65, 0.4) rectangle (1.45, -6.0);
    \fill[gray!4] (1.6, 0.4) rectangle (4.7, -6.0);
    \fill[colorOurs!5, draw=colorOurs!20, line width=0.5pt, rounded corners=4pt] (4.85, 0.4) rectangle (7.95, -6.0);

    \node[header] (h1) at (-0.1, 0) {DOM-Centric\\ \fontsize{6}{7}\selectfont (Structural)};
    \node[header] (h2) at (3.15, 0) {SoM-Centric\\ \fontsize{6}{7}\selectfont (Hybrid)};
    \node[header, text=colorOurs] (h3) at (6.4, 0) {\textsc{Avenir-Web}\\ \fontsize{6}{7}\selectfont (Direct Visual)};

    \node[card] (p1) at (-0.1, -1.5) {
        \begin{tikzpicture}[scale=0.55, every node/.style={transform shape}]
            \node[font=\sffamily\fontsize{11}{12}\selectfont, align=left, text=gray!90] at (0,0) {
                1: [Button] 'Submit'\\
                2: [Input] 'Search'\\
                3: [Link] 'Home'\\
                4: [Img] 'Logo'
            };
        \end{tikzpicture}\\[1pt]
        \fontsize{7.5}{8.5}\selectfont Cleaned DOM
    };
    \node[card] (p2) at (3.15, -1.5) {
        \begin{tikzpicture}[scale=0.45, every node/.style={transform shape}]
            \begin{scope}[line width=1.2pt, line join=miter]
                \draw[cyan] (0.1, 1.2) rectangle (0.8, 1.8);
                \node[text=cyan, font=\tiny, anchor=center] at (0.45, 1.5) {1};
                
                \draw[red] (0.1, 0.2) rectangle (2.2, 0.9);
                \node[text=red, font=\tiny, anchor=center] at (1.15, 0.55) {2};
                
                \draw[blue!70] (1.3, 1.1) rectangle (2.4, 1.7);
                \node[text=blue!70, font=\tiny, anchor=center] at (1.85, 1.4) {3};
            \end{scope}
        \end{tikzpicture}\\[1pt]
        \fontsize{7.5}{8.5}\selectfont Set-of-Mark
    };
    \node[our_card] (p3) at (6.4, -1.5) {
        \begin{tikzpicture}[scale=0.6, every node/.style={transform shape}]
            \fill[gray!10] (0.1, 1.55) rectangle (2.3, 1.7); 
            \fill[gray!10] (0.1, 0.1) rectangle (0.5, 1.45); 
            
            \foreach \x in {0.6, 1.2, 1.8} {
                \foreach \y in {0.1, 0.5, 0.9} {
                    \fill[gray!5, draw=gray!15, line join=miter] (\x, \y) rectangle (\x+0.5, \y+0.3);
                }
            }
        \end{tikzpicture}\\[1pt]
        \fontsize{7.5}{8.5}\selectfont Visual Scene
    };

    \node[card, minimum height=0.8cm] (r1) at (-0.1, -3.0) {Selector Mapping};
    \node[card, minimum height=0.8cm] (r2) at (3.15, -3.0) {Tag Association};
    \node[our_card, minimum height=0.8cm] (r3) at (6.4, -3.0) {GUI Grounding};

    \node[code_box, minimum height=1.0cm] (a1) at (-0.1, -4.5) {DOM-based\\ Action};
    \node[code_box, minimum height=1.0cm] (a2) at (3.15, -4.5) {Tag-based\\ Action};
    \node[our_code_box, minimum height=1.0cm] (a3) at (6.4, -4.5) {Mixture of Grounding\\ Experts (MoGE)};

    \draw[flow] (p1) -- (r1); \draw[flow] (r1) -- (a1);
    \draw[flow] (p2) -- (r2); \draw[flow] (r2) -- (a2);
    \draw[our_flow] (p3) -- (r3); \draw[our_flow] (r3) -- (a3);

\end{tikzpicture}
}
\caption{Architectural comparison of web agent grounding paradigms. DOM-Centric models rely on structural parsing of the page hierarchy. SoM-Centric systems utilize visual tagging to bridge the pixel-element gap. \textsc{Avenir-Web} uses MoGE for semantic-first grounding with hierarchical fallbacks, leveraging structured metadata to resolve complex interactions while maintaining direct visual grounding capabilities.}
\label{fig:comparison}
\end{figure}
\subsection{Mixture of Grounding Experts (MoGE)}\label{sec:moge}
The core operative module of \textsc{Avenir-Web} is the Mixture of Grounding Experts (MoGE), which prioritizes direct visual grounding using general-purpose multimodal models such as \textsc{Qwen3-VL}\cite{bai2025qwen3vltechnicalreport} and \textsc{Gemini 3 Pro}\cite{google_deepmind_gemini3pro}. This design emulates human interaction by treating the entire viewport, including nested iframes, as a unified visual canvas, thereby bypassing the structural complexity of underlying HTML code. By focusing on coordinate-based interaction, an approach increasingly adopted by native GUI agents~\cite{lin2024showuivisionlanguageactionmodelgui, qin2025uitarspioneeringautomatedgui, hong2024cogagentvisuallanguagemodel}, MoGE naturally resolves elements within iframes that frequently hinder DOM-centric agents. In edge cases where visual cues are insufficient or fine-grained structural control is required, the system falls back to semantic structural reasoning to ensure reliable grounding across diverse web paradigms~\cite{zheng2024seeact, gou2025uground, yan2023gpt4vwonderlandlargemultimodal} (see Figure~\ref{fig:seeact_noop}). Crucially, MoGE functions as a one-step grounding system for standard interactions, producing an executable action with a single model inference. To improve precision, it incorporates a visual annotation layer inspired by Set-of-Mark (SoM) prompting~\cite{yang2023setofmarkpromptingunleashesextraordinary}, overlaying interactive elements with unique identifiers to disambiguate closely spaced controls. This design enables less than one model call per step on average, streamlining the interaction loop while avoiding spatial context loss during HTML parsing~\cite{lu2024omniparserpurevisionbased, hsieh2025zonui3blightweightvisionlanguagemodel}. For more complex grounding scenarios, MoGE can be extended with iterative reasoning steps, following the \textit{Chain-of-Ground} paradigm~\cite{li2025chainofgroundimprovingguigrounding, lei2025textscguispotlightadaptiveiterativefocus}.

For specialized interactions, MoGE adopts a hierarchical decision-making architecture that prioritizes execution efficiency while maintaining robustness through progressively richer grounding strategies. The framework supports heterogeneous modalities via a tiered design in which point-based actions emphasize high-precision spatial localization in a normalized coordinate space, while preserving parallel semantic context to enable structural recovery when visual grounding fails to induce a valid state transition. Text entry follows a staged protocol that first attempts direct coordinate-based input via a virtual keyboard; upon failure or verification inconsistency, the system falls back to structural element targeting, and finally to a global search over candidate input fields ranked by spatial proximity or semantic similarity, with an LLM resolving remaining ambiguities. Dropdown interactions use a complementary fallback strategy, beginning with direct value assignment through script-level manipulation to handle system-native components that are often inaccessible to coordinate-based clicking; if unsuccessful, the system performs a semantic search over selectable elements and applies LLM-based reasoning to identify and select the most appropriate option.

\subsection{Adaptive Memory}

Sequential decision-making in long-horizon web tasks requires a memory mechanism that can balance detailed execution history with long-term strategic awareness. Without such a mechanism, agents often operate in a purely reactive manner, taking one step at a time without analyzing why previous actions may have failed to reach a goal~\cite{yao2023reactsynergizingreasoningacting, shinn2023reflexion}. This often results in repetitive failure loops where the system repeats the same unsuccessful action until reaching a step limit (see Figure~\ref{fig:memory_comparison}).

\begin{figure}[t]
\centering
\begin{tikzpicture}[
    avenir_font,
    scale=0.82, transform shape,
    node distance=0.1cm,
    panel_group/.style={rectangle, draw=gray!45, dashed, inner sep=4pt, rounded corners=6pt, fill=gray!8, minimum width=2.8cm, minimum height=6.2cm},
    hero_group/.style={panel_group, fill=orange!2, draw=orange!20},
    box/.style={rectangle, draw=black!60, rounded corners=2pt, align=center, fill=white, inner sep=2pt, thick, font=\sffamily\fontsize{6}{7}\selectfont},
    arch_module/.style={box, fill=blue!5, draw=blue!40, minimum width=2.4cm, minimum height=0.55cm},
    lost_node/.style={arch_module, dashed, draw=gray!40, fill=white, text=gray!50},
    window_frame/.style={draw=colorOurs, line width=1pt, rounded corners=4pt, fill=colorOurs!5, fill opacity=0.1},
    flow_arrow/.style={-stealth, draw=colorOurs, line width=0.8pt, shorten >=1pt, shorten <=1pt}
]

    \node[panel_group] (p1) {};
    \node[panel_group, right=0.25cm of p1] (p2) {};
    \node[hero_group, right=0.25cm of p2] (p3) {};

    \node[anchor=north, yshift=-4pt, font=\sffamily\tiny, text=gray!85] at (p1.north) {Full Context};
    \node[anchor=north, yshift=-4pt, font=\sffamily\tiny, text=gray!85] at (p2.north) {Fixed Window ($W=5$)};
    \node[anchor=north, yshift=-4pt, font=\sffamily\tiny\bfseries, text=colorOurs!90!black] at (p3.north) {Adaptive Memory (Ours)};

    \node[arch_module, below=0.55cm of p1.north] (n1) {Step 1};
    \node[arch_module, below=0.08cm of n1] (n2) {Step 2};
    \node[arch_module, below=0.08cm of n2] (n3) {Step 3};
    \node[arch_module, below=0.08cm of n3] (n4) {Step 4};
    \node[arch_module, below=0.08cm of n4] (n5) {Step 5};
    \node[arch_module, below=0.08cm of n5] (n6) {Step 6};
    \node[arch_module, below=0.08cm of n6] (n7) {Step 7};
    \node[anchor=south, yshift=2pt, font=\sffamily\fontsize{5}{6}\selectfont, text=red!70, fill=red!5, draw=red!20, rounded corners=2pt] at (p1.south) {Hallucination Risk};

    \node[lost_node, below=0.55cm of p2.north] (w1) {Step 1 (Lost)};
    \node[lost_node, below=0.08cm of w1] (w2) {Step 2 (Lost)};
    \node[arch_module, below=0.08cm of w2] (w3) {Step 3};
    \node[arch_module, below=0.08cm of w3] (w4) {Step 4};
    \node[arch_module, below=0.08cm of w4] (w5) {Step 5};
    \node[arch_module, below=0.08cm of w5] (w6) {Step 6};
    \node[arch_module, below=0.08cm of w6] (w7) {Step 7};
    \node[window_frame, fit=(w3) (w7), inner sep=1.5pt] (frame_p2) {};
    \node[anchor=south, yshift=2pt, font=\sffamily\fontsize{5}{6}\selectfont, text=orange!80, fill=orange!5, draw=orange!20, rounded corners=2pt] at (p2.south) {Context Blindness};

    \node[draw=teal!60, fill=teal!5, thick, rounded corners=3pt, minimum width=2.6cm, minimum height=1.3cm, anchor=north] (sum_box) at ([yshift=-0.65cm]p3.north) {};
    \node[anchor=north, font=\sffamily\fontsize{5}{6}\selectfont, text=teal!90, yshift=-2pt] at (sum_box.north) {LLM Summary};
    \node[draw=teal!30, fill=white, rounded corners=2pt, minimum width=2.2cm, minimum height=0.6cm, align=center] at ([yshift=-0.1cm]sum_box.center) {
        \fontsize{5}{6}\selectfont Steps 1--10 Distilled
    };

    \node[window_frame, minimum width=2.6cm, minimum height=3.0cm, anchor=south, fill opacity=0.05] (win_bg) at ([yshift=1.2cm]p3.south) {};
    \node[anchor=north, font=\sffamily\fontsize{5}{6}\selectfont, text=colorOurs, yshift=-2pt] at (win_bg.north) {Window ($W=5$)};
    
    \node[arch_module, below=0.4cm of win_bg.north] (s6) {Step 6};
    \node[arch_module, below=0.05cm of s6] (s7) {Step 7};
    \node[arch_module, dashed, draw=gray!40, fill=white, text=gray!45, below=0.1cm of s7, minimum height=1.0cm] (future) {Residual Window\\Capacity};

    \node[anchor=south, yshift=2pt, font=\sffamily\fontsize{5}{6}\selectfont, text=colorOurs!90!black, fill=colorOurs!10, draw=colorOurs!30, rounded corners=2pt] at (p3.south) {Strategic Awareness};

\end{tikzpicture}
\vspace{2pt}
\caption{Comparison of memory architectures for long-horizon web navigation. While \textit{Full Context} risks hallucination due to excessive tokens and \textit{Fixed Window} suffers from context blindness as early steps are lost, our \textbf{Adaptive Memory} maintains strategic awareness by combining a distilled recursive summary of past actions with a precise sliding window of recent interactions.}\label{fig:memory_comparison}
\end{figure}

To address the propensity for navigational drift and repetitive errors, we introduce Adaptive Memory, which balances tactical interaction history with strategic awareness via Chunked Recursive Summarization. This mechanism is inspired by the need for long-term reasoning and self-correction in complex agentic workflows~\cite{yao2023reactsynergizingreasoningacting, shinn2023reflexion}. The system operates over a sliding window of size $W$ (default $W=5$), periodically distilling its raw interaction buffer $\mathcal{B}_k$ into a more abstract, persistent memory state $\mathcal{M}_k$. The buffer for the $k$-th chunk is defined as:
\begin{equation}
    \mathcal{B}_k = \{(a_i, o_i, S_i)\}_{i=(k-1)W+1}^{kW}
\end{equation}
This recursive distillation preserves high-level strategic awareness while preventing the context-induced hallucination~\cite{Huang_2025} that often occurs when an agent is overwhelmed by a long and repetitive history of low-level actions. Upon the completion of each interaction chunk, the memory state is updated as:
\begin{equation}
    \mathcal{M}_k = \mathcal{G}_{\phi}(\mathcal{M}_{k-1}, \mathcal{B}_k, \mathcal{E}_k)
\end{equation}
where $\mathcal{G}_{\phi}$ is the distillation function parameterized by $\phi$, and $\mathcal{E}_k$ represents the failure reflection buffer containing distilled traces of execution errors within the chunk. This ensures that the agent retains long-term situational awareness without context overflow. Furthermore, to ensure that critical failures are never lost during summarization, any execution error or unexpected feedback is immediately summarized by the LLM and added to the summary buffer. This \textit{Failure Reflection} allows the agent to reason over past errors with distilled clarity, even after the corresponding step has been summarized or removed from the active window.

To facilitate this reflection, the system implements a robust outcome detection mechanism that verifies the impact of every action. It compares the page state before and after execution to determine whether an interaction has resulted in a meaningful transition. The mechanism evaluates state changes through a prioritized hierarchy, looking for changes in visible text, interactive elements, focus, the URL, scroll position, or modal popups. For key interactions such as point-based clicking or textual input, the absence of detected changes triggers a failure flag. When a failure is confirmed, the system logs a warning and captures a coordinate-annotated screenshot to assist in future reasoning, while propagating the failure status to the Task-Tracking Checklist to prompt a strategic retry.

%% file: sections/petfinder_example.tex
\begin{figure}[t]
\centering
\resizebox{0.9\linewidth}{!}{
\begin{tikzpicture}[
    avenir_font,
    ui_frame/.style={rectangle, draw=colorStroke, fill=white, rounded corners=4pt, line width=0.4pt, drop shadow={opacity=0.08, shadow xshift=1.5pt, shadow yshift=-1.5pt}},
    url_bar/.style={rectangle, fill=colorBG, draw=colorStroke, rounded corners=2pt, minimum height=0.4cm, font=\sffamily\tiny, text=colorTextMuted, align=center, inner sep=2pt},
    agent_card/.style={rectangle, draw=colorOurs!30, fill=colorOurs!2, rounded corners=3pt, inner sep=4pt},
    process_bubble/.style={rectangle, draw=colorStroke, fill=white, rounded corners=2pt, text width=6.6cm, align=left, inner sep=4pt, font=\sffamily\scriptsize, drop shadow={opacity=0.03}},
    status_tag/.style={rectangle, rounded corners=1.5pt, inner sep=2pt, font=\sffamily\fontsize{5}{6}\selectfont\bfseries, minimum width=34pt, align=center},
    tag_success/.style={status_tag, fill=green!15, text=green!60!black},
    tag_failed/.style={status_tag, fill=red!10, text=red!80},
    tag_pending/.style={status_tag, fill=colorTextMuted!10, text=colorTextMuted},
    checklist_row/.style={rectangle, fill=white, draw=colorGrid, rounded corners=1.5pt, minimum width=6.8cm, text width=6.4cm, minimum height=0.55cm, align=left, inner sep=3pt, font=\sffamily\scriptsize},
    column_bg/.style={rectangle, rounded corners=3pt, inner sep=4pt}
]

    \node [ui_frame, minimum width=8.2cm, minimum height=7.8cm] (frame) at (0,0) {};
    
    \fill [colorGrid!30, rounded corners=4pt] ([yshift=-0.6cm]frame.north west) rectangle (frame.north east);
    
    \fill[red!30] ([xshift=10pt, yshift=-8.5pt]frame.north west) circle (1.5pt);
    \fill[orange!30] ([xshift=17pt, yshift=-8.5pt]frame.north west) circle (1.5pt);
    \fill[green!30] ([xshift=24pt, yshift=-8.5pt]frame.north west) circle (1.5pt);

    \node [url_bar, minimum width=5.8cm] (url) at ([yshift=-8.5pt]frame.north) {
        \texttt{https://petfinder.com/}
    };

    \node [agent_card, minimum width=7.8cm, text width=7.4cm] (task_header) at ([yshift=-1.1cm]frame.north) {
        \textbf{Instruction:} Find cats available for adoption within 10 miles of zip code 94587, Young or adult-age cats, sorted by Oldest Addition.
    };

    
    \node [column_bg, fill=colorOurs!4, minimum width=7.8cm, minimum height=3.1cm, anchor=north] (eip_panel) at ([yshift=-1.6cm]frame.north) {};
    \node [anchor=north west, font=\sffamily\bfseries\tiny, text=colorOurs] (eip_label) at ([xshift=6pt, yshift=-4pt]eip_panel.north west) {EXPERIENCE-IMITATION PLANNER};
    
    \node [process_bubble] (s1) at ([yshift=-1.0cm]eip_panel.north) {
        Click \textit{"Find a cat"} to start the cat adoption search.
    };
    \node [process_bubble, below=0.08cm of s1] (s2) {
        Enter \textit{"94587"} in location and set distance to \textit{10 miles}.
    };
    \node [process_bubble, below=0.08cm of s2] (s3) {
        Apply filters by selecting \textit{Young} and \textit{Adult} options.
    };
    \node [process_bubble, below=0.08cm of s3] (s4) {
        Change sort order to \textit{Oldest Addition} using dropdown.
    };

    \node [column_bg, fill=gray!4, minimum width=7.8cm, minimum height=2.7cm, anchor=north] (ttc_panel) at ([yshift=-0.1cm]eip_panel.south) {};
    \node [anchor=north west, font=\sffamily\bfseries\tiny, text=colorOurs] (ttc_label) at ([xshift=6pt, yshift=-4pt]ttc_panel.north west) {TASK-TRACKING CHECKLIST};

    \node [checklist_row] (c1) at ([yshift=-0.8cm]ttc_panel.north) {
        Cats within 10 miles of 94587
    };
    \node [tag_success, anchor=east] at ([xshift=-4pt]c1.east) {SUCCESS};
    
    \node [checklist_row, below=0.1cm of c1] (c2) {
        Young or adult age cats only
    };
    \node [tag_pending, anchor=east] at ([xshift=-4pt]c2.east) {PENDING};

    \node [checklist_row, below=0.1cm of c2] (c3) {
        Sorted by oldest addition
    };
    \node [tag_failed, anchor=east] at ([xshift=-4pt]c3.east) {FAILED};

\end{tikzpicture}}
\vspace{8pt}
\caption{\textsc{Avenir-Web} execution context for \url{petfinder.com}. The figure illustrates the integration of Experience-Imitation Planning (EIP) and Task-Tracking Checklist with status feedback.}\label{fig:petfinder_example}
\end{figure}

%% file: sections/experiments.tex
\section{Experiments}\label{sec:experiments}
\begin{table*}[t]
\centering
\caption{Task Success Rate on \textsc{Online-Mind2Web}~\cite{xue2025an} across difficulty levels. Results for \textsc{Avenir-Web} (the new open-source state-of-the-art on \textsc{Online-Mind2Web}) are shown alongside industry baselines and specialized proprietary agents.}\label{tab:main_results}
\vspace{3pt}
\small
\resizebox{\textwidth}{!}{
\begin{tabular}{llcccccc}
\toprule
Agent & Main Model & Provider & Open Source & Easy & Med. & Hard & Overall \\
\midrule
\rowcolor{colorGrid} \multicolumn{8}{l}{\textbf{\textcolor{colorTextMuted}{Proprietary Models}}} \\
\addlinespace[2pt]
\textcolor{colorTextMuted}{Navigator~\cite{navigator}} & \textcolor{colorTextMuted}{\texttt{n1-preview-11-2025}} & \textcolor{colorTextMuted}{Yutori} & \textcolor{colorTextMuted}{\ding{55}} & \textcolor{colorTextMuted}{84.0} & \textcolor{colorTextMuted}{62.2} & \textcolor{colorTextMuted}{48.7} & \textcolor{colorTextMuted}{64.7} \\
\textcolor{colorTextMuted}{Operator~\cite{openai2025introducing_operator}} & \textcolor{colorTextMuted}{\texttt{OpenAI Computer-Using Agent}} & \textcolor{colorTextMuted}{OpenAI} & \textcolor{colorTextMuted}{\ding{55}} & \textcolor{colorTextMuted}{73.5} & \textcolor{colorTextMuted}{59.4} & \textcolor{colorTextMuted}{39.2} & \textcolor{colorTextMuted}{58.3} \\
\textcolor{colorTextMuted}{Google Computer Use~\cite{gemini_computer_use}} & \textcolor{colorTextMuted}{\texttt{Gemini 2.5 Computer Use}} & \textcolor{colorTextMuted}{Google DeepMind} & \textcolor{colorTextMuted}{\ding{55}} & \textcolor{colorTextMuted}{77.1} & \textcolor{colorTextMuted}{55.2} & \textcolor{colorTextMuted}{45.9} & \textcolor{colorTextMuted}{57.3} \\
\textcolor{colorTextMuted}{ACT-1-20250814~\cite{enhans}} & \textcolor{colorTextMuted}{\texttt{o3-2025-04-16} and \texttt{Claude-sonnet-4-20250514}} & \textcolor{colorTextMuted}{Enhans} & \textcolor{colorTextMuted}{\ding{55}} & \textcolor{colorTextMuted}{71.1} & \textcolor{colorTextMuted}{52.4} & \textcolor{colorTextMuted}{32.4} & \textcolor{colorTextMuted}{52.7} \\
\textcolor{colorTextMuted}{Claude Computer Use 3.7~\cite{anthropic2025sonnet45}} & \textcolor{colorTextMuted}{\texttt{Claude-3.7-sonnet-20250219}} & \textcolor{colorTextMuted}{Anthropic} & \textcolor{colorTextMuted}{\ding{55}} & \textcolor{colorTextMuted}{75.9} & \textcolor{colorTextMuted}{41.3} & \textcolor{colorTextMuted}{27.0} & \textcolor{colorTextMuted}{47.3} \\
\textcolor{colorTextMuted}{ACT-1-20250703~\cite{enhans}} & \textcolor{colorTextMuted}{\texttt{o3-2025-04-16} and \texttt{Claude-sonnet-4-20250514}} & \textcolor{colorTextMuted}{Enhans} & \textcolor{colorTextMuted}{\ding{55}} & \textcolor{colorTextMuted}{53.7} & \textcolor{colorTextMuted}{39.2} & \textcolor{colorTextMuted}{24.3} & \textcolor{colorTextMuted}{39.5} \\
\textcolor{colorTextMuted}{Claude 3.5~\cite{anthropic2024claude35}} & \textcolor{colorTextMuted}{\texttt{Claude-3-5-sonnet-20241022}} & \textcolor{colorTextMuted}{Anthropic} & \textcolor{colorTextMuted}{\ding{55}} & \textcolor{colorTextMuted}{51.8} & \textcolor{colorTextMuted}{16.1} & \textcolor{colorTextMuted}{8.1} & \textcolor{colorTextMuted}{24.0} \\
\midrule
\rowcolor{colorGrid} \multicolumn{8}{l}{\textbf{Open-Source Baselines}} \\
\addlinespace[2pt]
SeeAct~\cite{zheng2024seeact} & \texttt{gpt-4o-2024-08-06} & OSU & \ding{51} & 51.8 & 28.0 & 9.5 & 30.0 \\
Agent-E~\cite{abuelsaad2024agente} & \texttt{gpt-4o-2024-08-06} & Emergence AI & \ding{51} & 51.8 & 23.1 & 6.8 & 27.0 \\
Browser Use~\cite{browseruse} & \texttt{gpt-4o-2024-08-06} & Browser Use & \ding{51} & 44.6 & 23.1 & 10.8 & 26.0 \\
\midrule
\rowcolor{colorOurs!15} \multicolumn{8}{l}{\textbf{Avenir-Web (Ours)}} \\
\addlinespace[2pt]
Avenir-Web & Gemini 3 Pro & Ours & \ding{51} & 74.1 & 54.6 & 30.3 & \textbf{53.7} \\
Avenir-Web & Qwen-3-VL-8B & Ours & \ding{51} & 42.0 & 23.8 & 11.8 & 25.7 \\
\bottomrule
\end{tabular}
}
\end{table*}

\subsection{Experimental Setup}
\textbf{Benchmark.} We evaluate \textsc{Avenir-Web} on \textsc{Online-Mind2Web}~\cite{xue2025an}, an online evaluation benchmark consisting of 300 diverse and realistic tasks spanning 136 websites. This environment enables the evaluation of web agents under a setting that approximates real-world user interaction, encompassing dynamic content, complex Document Object Model (DOM) structures, and state-dependent workflows.

\textbf{Metrics.} We report the \textit{Task Success Rate (TSR)}, which measures the percentage of tasks where the agent successfully reaches the target goal state. To facilitate scalable evaluation, we utilize a novel \textit{LLM-as-a-Judge} automatic evaluation method. This approach, powered by \textit{o4-mini}, achieves an average agreement rate of 85.7\% with human judgment and maintains a narrow success rate gap of just 3.8\%, which is substantially higher than existing automated methods.

\textbf{Models \& Baselines.} We benchmark \textsc{Avenir-Web} against the primary open-source baselines \textsc{SeeAct}~\cite{zheng2024seeact}, \textsc{Browser Use}~\cite{browseruse}, and \textsc{Agent-E}, alongside state-of-the-art proprietary agents such as \textsc{Operator}~\cite{openai2025introducing_operator}, \textsc{Gemini 2.5 Computer Use}~\cite{gemini_computer_use}, and \textsc{ACT-1}~\cite{enhans}. We utilize Gemini 3 Pro~\cite{google_deepmind_gemini3pro} as our primary action backbone. Furthermore, our architecture leverages specialized models for auxiliary reasoning: the Experience-Imitation Planning (EIP) module employs \textit{Claude 4.5 Sonnet}~\cite{anthropic2025sonnet45} for strategic roadmap synthesis, while the Task-Tracking Checklist is managed by a lightweight \textit{Qwen-3-VL-8B}~\cite{bai2025qwen3vltechnicalreport} model to ensure responsive state monitoring.

\subsection{Main Results}
Table~\ref{tab:main_results} presents a comprehensive comparison on the \textsc{Online-Mind2Web} benchmark. \textsc{Avenir-Web} achieves a new open-source state-of-the-art performance on \textsc{Online-Mind2Web}, significantly outperforming existing baselines such as \textsc{SeeAct}~\cite{zheng2024seeact} (30.0\%), \textsc{Agent-E} (27.0\%), and \textsc{Browser Use}~\cite{browseruse} (26.0\%). 

Our primary configuration, \textsc{Avenir-Web} (Gemini 3 Pro), achieves a success rate of \textbf{53.7\%}, representing an absolute improvement of at least \textbf{23.7\%} over prior open-source systems. Notably, we also provide a fully open-source configuration using \textit{Qwen-3-VL-8B} as the main action model. This version achieves a success rate of \textbf{25.7\%}, demonstrating that our framework enables even a lightweight 8B model to reach performance levels comparable to existing open-source baselines that rely on much larger proprietary models. While it trails specialized commercial systems like \href{https://yutori.com/blog/introducing-navigator}{Yutori Navigator}~\cite{navigator} (64.7\%), OpenAI Operator (58.3\%)~\cite{openai2025introducing_operator} and Google Gemini 2.5 Computer Use (57.3\%)~\cite{gemini_computer_use}, it outperforms other major proprietary baselines including \texttt{ACT-1-20250814}~\cite{enhans} (52.7\%) and Claude Computer Use 3.7~\cite{anthropic2025sonnet45} (47.3\%). This establishes \textsc{Avenir-Web} as the new open-source state-of-the-art on \textsc{Online-Mind2Web}, effectively bridging the performance gap with proprietary frontier models.

\subsection{Ablation Studies}\label{sec:ablation_studies}

To evaluate the impact of the four core components of \textsc{Avenir-Web}, namely Experience-Imitation Planning (EIP), Mixture of Grounding Experts (MoGE), Task-Tracking Checklist, and Adaptive Memory, we perform ablation studies on a 50-tasks subset of Online-Mind2Web(Table~\ref{tab:ablation_grounding}).

Removing \textit{Experience-Imitation Planning} drops the success rate from 48.0\% to 36.0\%, while disabling the \textit{Task-Tracking Checklist} leads to a 4.0\% decline, confirming the necessity of site-specific knowledge and verifiable state tracking. Disabling \textit{MoGE} reduces performance to 40.0\% due to struggles with fine-grained elements that visual-only models often misinterpret. Finally, replacing recursive \textit{Adaptive Memory} with simple sliding windows ($W=5$ or $W=\infty$) decreases performance; specifically, the $W=\infty$ setting causes significant hallucination as the context window saturates, validating that recursive distillation preserves situational awareness while preventing context-induced errors in long-horizon trajectories.
\subsection{Qualitative Analysis}\label{sec:qualitative_analysis}
To further investigate the impact of the proposed grounding and memory mechanisms, we perform a qualitative trajectory analysis and provide representative screenshot-based trajectories in Appendix~\ref{sec:appendix_case_study}. As shown in Figure~\ref{fig:seeact_noop}, \textsc{Avenir-Web} maintains stable progress through multi-step interfaces where DOM-centric baselines stall on iframe-heavy layouts, highlighting the practical benefits of MoGE grounding and memory-aware task tracking in real web settings.

\begin{table}[!t]
\centering
\caption{Ablation study on MoGE, EIP, and Adaptive Memory on a Online-Mind2Web subset of 50 tasks (Backbone: Gemini 3 Flash).}\label{tab:ablation_grounding}
\small
\resizebox{\columnwidth}{!}{
\begin{tabular}{lc}
\toprule
Configuration & Success Rate (\%) \\
\midrule
Full Model (\textsc{Avenir-Web}) & \textbf{48.0} \\
\quad w/o Task-Tracking Checklist & 44.0 \\
\quad w/o Adaptive Memory ($W=5$) & 42.0 \\
\quad w/o Adaptive Memory ($W=\infty$) & 36.0 \\
\quad w/o Experience-Imitation Planning (EIP) & 36.0 \\
\quad w/o Mixture of Grounding Expert (MoGE) & 40.0 \\
\bottomrule
\end{tabular}
}
\end{table}
\vspace{-12pt}

%% file: sections/conclusion.tex
\section{Conclusion}

In this work, we presented \textsc{Avenir-Web}, a new open-source state-of-the-art web agent on \textsc{Online-Mind2Web} designed to overcome the reliability bottlenecks inherent in existing web agents on live websites. By unifying \textit{Experience-Imitation Planning} to incorporate site-specific procedural knowledge, a \textit{Mixture of Grounding Experts (MoGE)} for precise cross-modal element interaction, and a \textit{Task-Tracking Checklist} with \textit{Adaptive Memory} for resilient state management, \textsc{Avenir-Web} effectively bridges the gap between high-level user intent and low-level execution on dynamic web interfaces.

Our extensive evaluation on the \textsc{Online-Mind2Web}~\cite{xue2025an} benchmark demonstrates that \textsc{Avenir-Web} sets a new open-source state-of-the-art on \textsc{Online-Mind2Web}, achieving a \textbf{53.7\%} success rate and outperforming prior baselines by a significant margin while remaining competitive with top-tier proprietary models. Ablation results further confirm that each module contributes to reliability, with the largest drops observed when removing Experience-Imitation Planning or effective memory management (Table~\ref{tab:ablation_grounding}). These results validate the efficacy of our dual-layer reasoning architecture and hybrid grounding strategy in handling the complexity of modern web tasks. Furthermore, the integration of strategic planning and recursive memory distillation not only improves task success but also enhances operational efficiency by reducing redundant trial-and-error exploration and mitigating token-intensive context bloat. We posit that \textsc{Avenir-Web} provides a foundational step towards more autonomous, general-purpose digital assistants capable of navigating the open web with human-level reliability. Subsequent research will explore further optimization of agent latency and investigate the scalability of experience-guided planning across a broader range of digital applications.
 
 \textbf{Social Impact and Limitations.} While \textsc{Avenir-Web} enhances productivity by automating routine workflows, its deployment involves critical safety and ethical considerations, including privacy risks and the potential for harmful actions. Technical limitations such as grounding accuracy and latency also persist.

%% file: sections/appendix.tex
\clearpage
\appendix
\onecolumn

\begin{center}
    {\LARGE \bf Appendix: Avenir-Web Supplemental Material}
\end{center}
\vspace{10pt}

\section{Case Study and Qualitative Analysis}
\label{sec:appendix_case_study}

\begin{figure}[h!]
\centering
\resizebox{0.55\columnwidth}{!}{
\begin{tikzpicture}[
    avenir_font,
    node distance=0.2cm,
    >=stealth
]
    \node[card, minimum width=12.0cm, minimum height=1.8cm] (instrBox) at (6.0, 0) {};
    \node[fill=colorOurs!15, minimum width=12.0cm, minimum height=0.5cm, anchor=north, rounded corners=2pt] (instrHeader) at (instrBox.north) {};
    \node[anchor=center, font=\sffamily\footnotesize\bfseries, text=colorOurs] at (instrHeader.center) {TASK SPECIFICATION};
    
    \node[anchor=north west, text width=5.5cm, font=\sffamily] (instrCol1) at ([yshift=-0.2cm, xshift=0.3cm]instrHeader.south west) {
        {\color{gray!90}\footnotesize INSTRUCTION}\\[1pt]
        {\footnotesize Open the reviews of a recipe with beef sirloin.}
    };
    \node[anchor=north west, text width=5.5cm, font=\sffamily] (instrCol2) at ([xshift=0.4cm]instrCol1.north east) {
        {\color{gray!90}\footnotesize TARGET URL}\\[1pt]
        {\footnotesize \urlstyle{same}\url{https://allrecipes.com}}
    };
    
    \node[avenir_bg, minimum width=12.0cm, minimum height=6.8cm, below=0.3cm of instrBox] (panelTop) {};
    \node[anchor=center] at (panelTop.center) {\includegraphics[width=11.6cm, height=6.5cm, keepaspectratio]{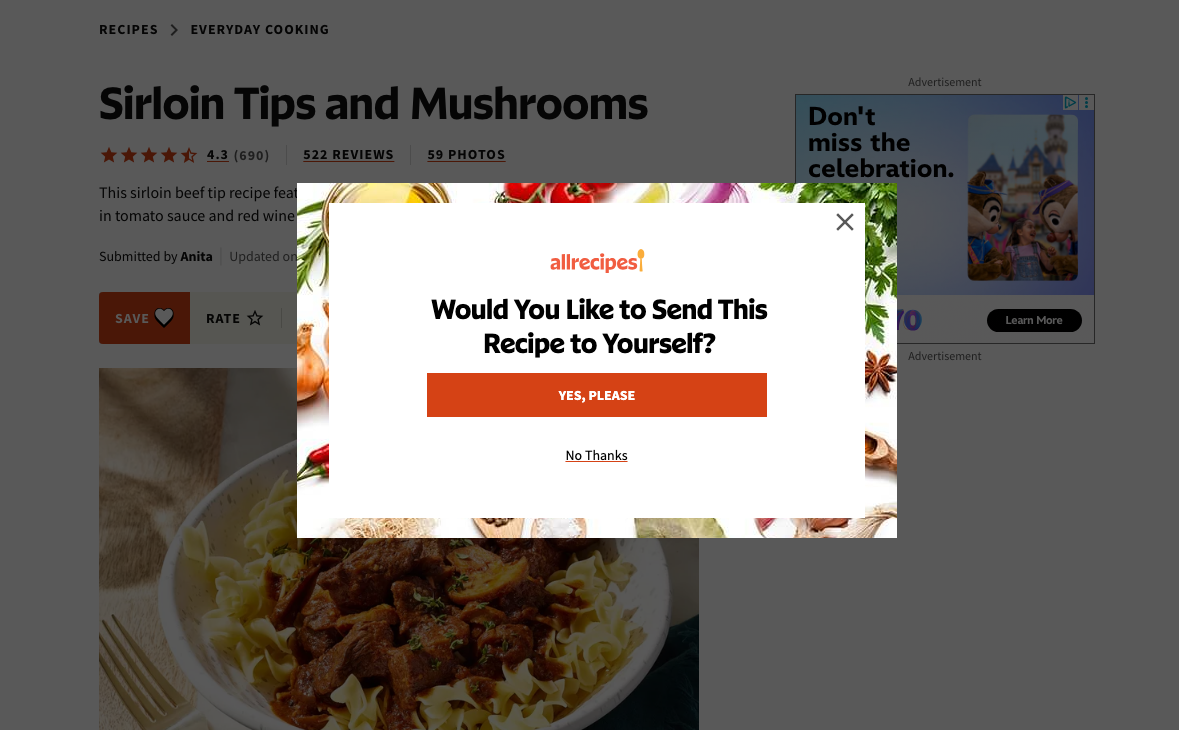}};

    \node[card, minimum width=5.8cm, minimum height=4.6cm, below=0.4cm of panelTop.south west, anchor=north west] (panelM) {};
    \node[fill=gray!10, minimum width=5.8cm, minimum height=0.6cm, anchor=north, rounded corners=2pt] (headerM) at (panelM.north) {};
    \node[anchor=center, font=\sffamily\scriptsize\bfseries, text=gray!80] at (headerM.center) {SeeAct (Failure)};
    
    \node[step_node, minimum width=5.3cm] (p1) at ([yshift=-1.1cm]panelM.north) {Type 'Beef Sirloin' into search};
    \node[step_node, minimum width=5.3cm, below=0.1cm of p1] (p2) {Press Enter};
    \node[step_node, minimum width=5.3cm, below=0.1cm of p2] (p3) {Click 'Sirloin Tips' result};
    \node[step_node, minimum width=5.3cm, below=0.1cm of p3, draw=colorStroke, fill=gray!8, text=colorTextMuted] (s1) {No Operation};
    \node[step_node, minimum width=5.3cm, below=0.1cm of s1, draw=colorStroke, fill=gray!8, text=colorTextMuted] (s2) {No Operation};
    \node[step_node, minimum width=5.3cm, below=0.1cm of s2, draw=red!60, fill=red!5, text=red!80] (s3) {Exceed Limit};
    
    \draw[modern_arrow] (p1) -- (p2);
    \draw[modern_arrow] (p2) -- (p3);
    \draw[modern_arrow] (p3) -- (s1);
    \draw[modern_arrow] (s1) -- (s2);
    \draw[modern_arrow] (s2) -- (s3);

    \node[card, minimum width=5.8cm, minimum height=4.6cm, below=0.4cm of panelTop.south east, anchor=north east, draw=colorOurs!40, fill=colorOurs!2] (panelR) {};
    \node[fill=colorOurs!15, minimum width=5.8cm, minimum height=0.6cm, anchor=north, rounded corners=2pt] (headerR) at (panelR.north) {};
    \node[anchor=center, font=\sffamily\scriptsize\bfseries, text=colorOurs] at (headerR.center) {Avenir-Web (Ours)};

    \node[step_node, minimum width=5.3cm, draw=colorOurs!40, fill=colorOurs!10, text=colorOurs!80] (o1) at ([yshift=-1.1cm]panelR.north) {Type 'Beef Sirloin' and press Enter};
    \node[step_node, minimum width=5.3cm, below=0.1cm of o1, draw=colorOurs!40, fill=colorOurs!10, text=colorOurs!80] (o2) {Click 'Sirloin Tips' result};
    \node[step_node, minimum width=5.3cm, below=0.1cm of o2, draw=green!80!black, fill=green!15, text=green!60!black] (o3) {Close modal via close button};
    \node[step_node, minimum width=5.3cm, draw=colorOurs!40, fill=colorOurs!10, text=colorOurs!80, below=0.1cm of o3] (o4) {Click reviews link};
    \node[step_node, minimum width=5.3cm, draw=colorOurs!40, fill=colorOurs!10, text=colorOurs!80, below=0.1cm of o4] (o5) {Task Success};

     \draw[modern_arrow_bold] (o1) -- (o2);
     \draw[modern_arrow_bold] (o2) -- (o3);
     \draw[modern_arrow_bold] (o3) -- (o4);
     \draw[modern_arrow_bold] (o4) -- (o5);

\end{tikzpicture}
}
\caption{Comparative trajectory analysis on \url{allrecipes.com}. The SeeAct baseline fails to operate within nested iframes, resulting in a sequence of non-responsive actions. In contrast, \textsc{Avenir-Web} successfully navigates the complex interface by leveraging the MoGE grounding module and modal recovery.}\label{fig:seeact_noop}
\end{figure}
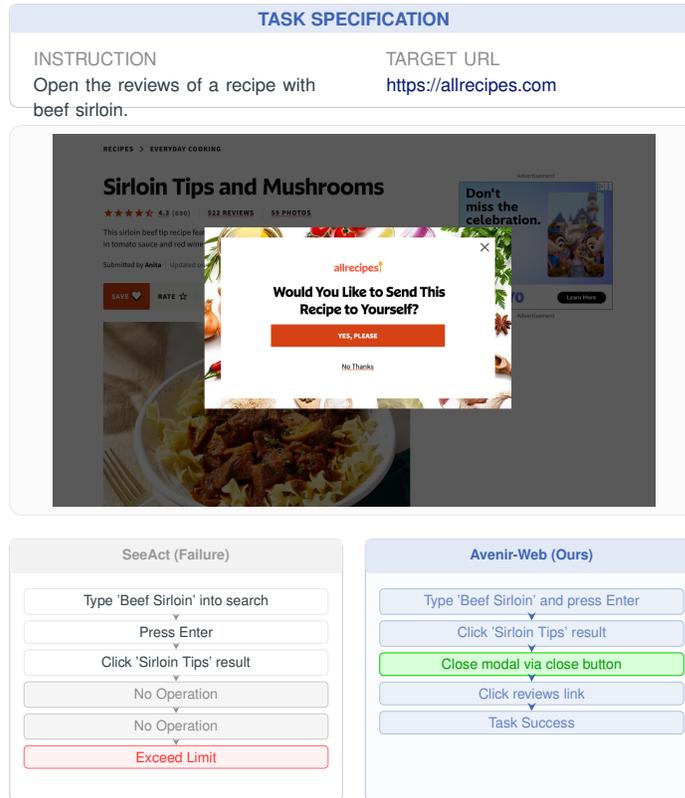

To further investigate the impact of the proposed grounding and memory mechanisms, we perform a qualitative trajectory analysis. Figure~\ref{fig:seeact_noop} illustrates a representative failure mode for the \textsc{SeeAct}~\cite{zheng2024seeact} baseline, which lacks the MoGE module. In this scenario, the baseline agent fails to navigate the interface due to its inability to operate within nested iframes, resulting in repeated non-responsive actions and eventual task timeout. Conversely, \textsc{Avenir-Web} demonstrates superior navigational resilience by synergizing visual coordinate prediction with semantic fallbacks. 

As shown in the success trajectory on \url{allrecipes.com} (Figure~\ref{fig:seeact_noop}), the agent effectively manages a multi-stage workflow involving: (i) an integrated text entry and search step where it types ``Beef Sirloin'' and immediately executes the submission, (ii) precise interaction to navigate to the recipe details, (iii) detection and dismissal of an obstructing modal window that appears prior to the target interaction, and (iv) successful execution of the review link interaction following modal recovery. This execution highlights how the MoGE module resolves spatial ambiguities, while the Adaptive Memory and Task-Tracking Checklist prevent the agent from entering the repetitive stalling loops characteristic of purely visual or DOM-centric baselines~\cite{gur2024a, he2024webvoyagerbuildingendtoendweb}.

\section{Detailed Case Study: Recreation.gov}
\label{sec:appendix_recreation_case}

To further illustrate the operational resilience of \textsc{Avenir-Web} on live websites, we provide a step-by-step execution trace for a task on \url{recreation.gov}. The metadata for this task is summarized below:

\begin{tcolorbox}[colback=colorBG, colframe=colorGrid, boxrule=0.5pt, sharp corners, fontupper=\sffamily\small]
\textbf{Task Instruction:} Check permit availability for a group of 4 in Brooks Camp, Katmai National Park for next Saturday. \\
\textbf{Target Website:} \url{https://www.recreation.gov/} \\
\textbf{Task ID:} \texttt{502e864440283214e0180645015f568b\_110325}
\end{tcolorbox}

\begin{figure}[h!]
\centering
\begin{subfigure}{0.23\textwidth}
    \includegraphics[width=\textwidth]{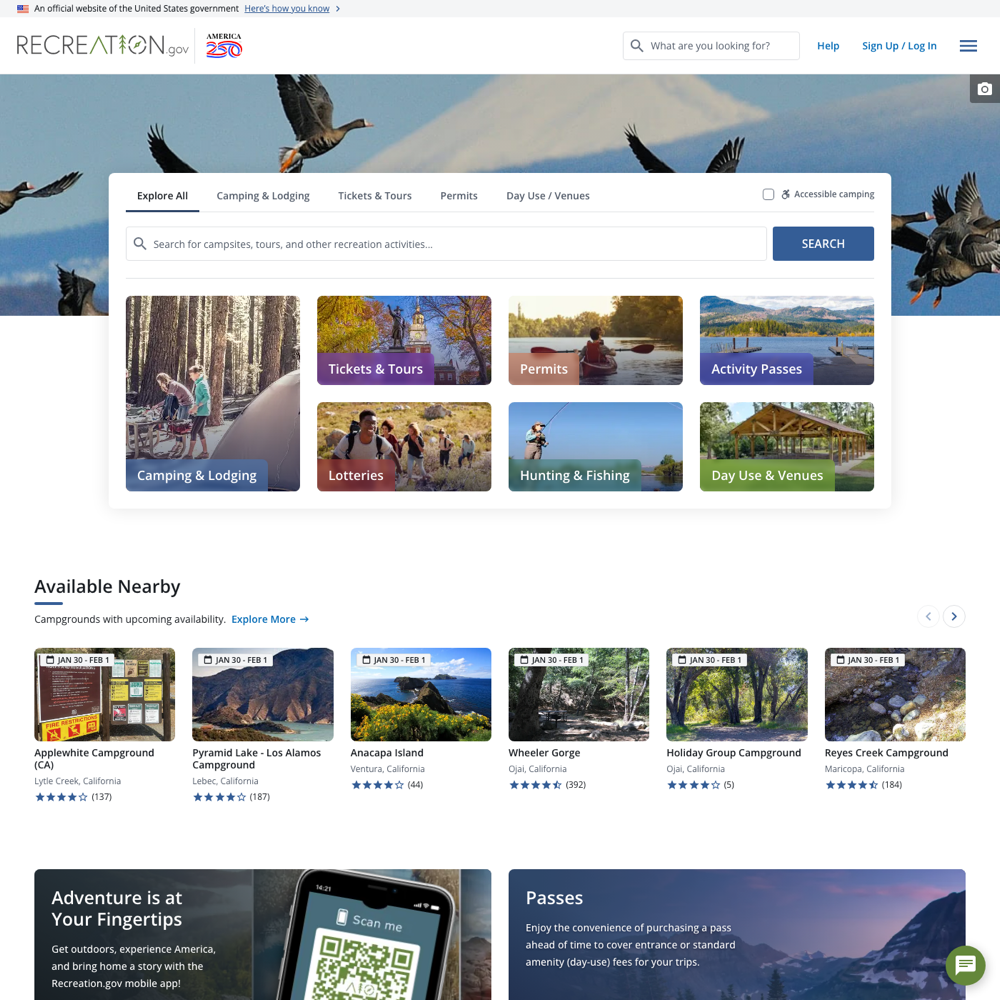}
    \caption{Step 1: \texttt{TYPE} ``Brooks Camp Katmai'' into the main search bar.}
\end{subfigure}
\hfill
\begin{subfigure}{0.23\textwidth}
    \includegraphics[width=\textwidth]{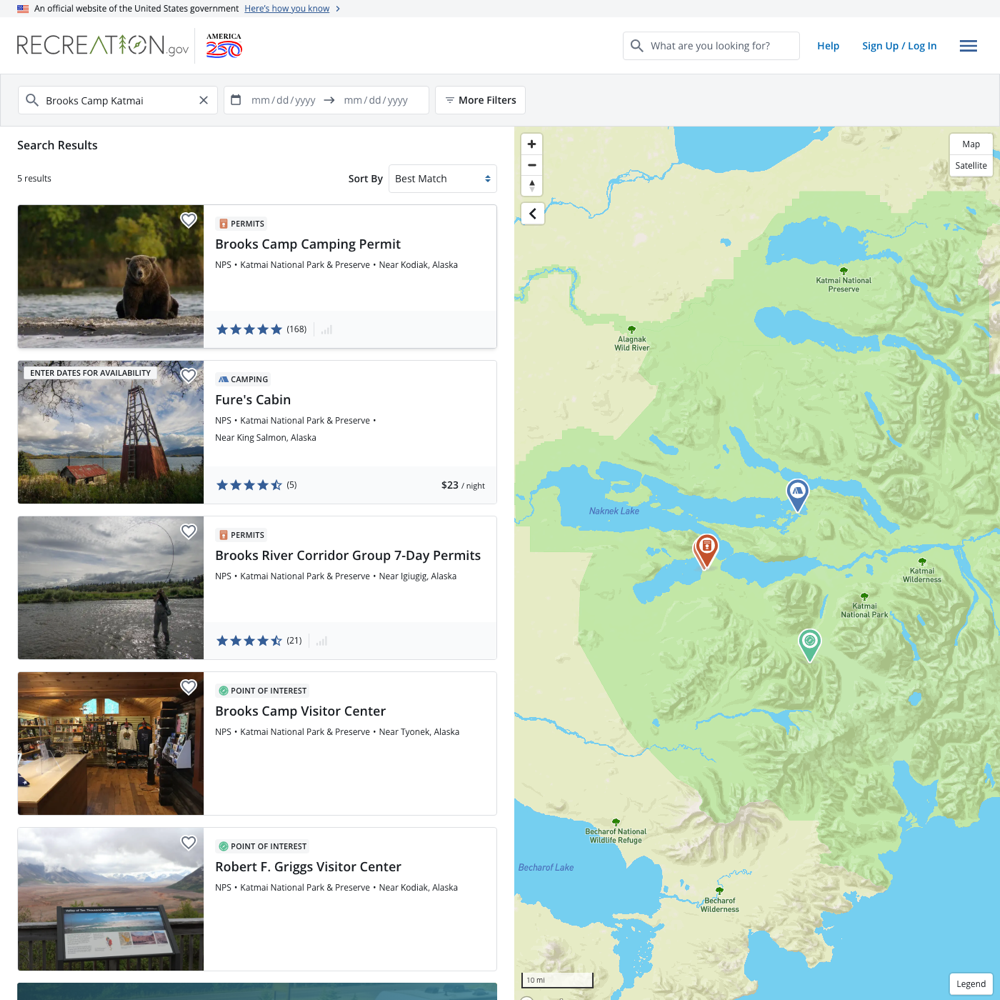}
    \caption{Step 2: \texttt{CLICK} the ``Brooks Camp Camping Permit'' result link.}
\end{subfigure}
\hfill
\begin{subfigure}{0.23\textwidth}
    \includegraphics[width=\textwidth]{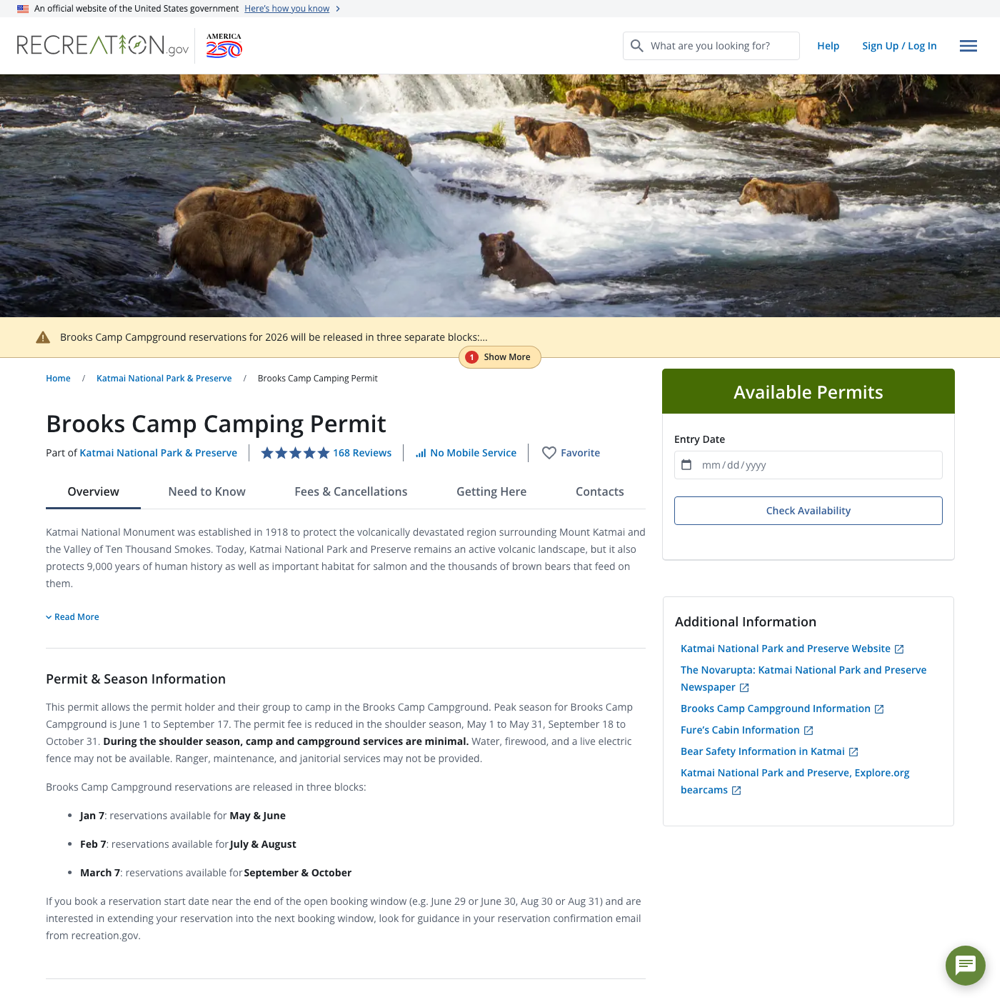}
    \caption{Step 3: \texttt{CLICK} the ``Entry Date'' input field to open the calendar picker.}
\end{subfigure}
\hfill
\begin{subfigure}{0.23\textwidth}
    \includegraphics[width=\textwidth]{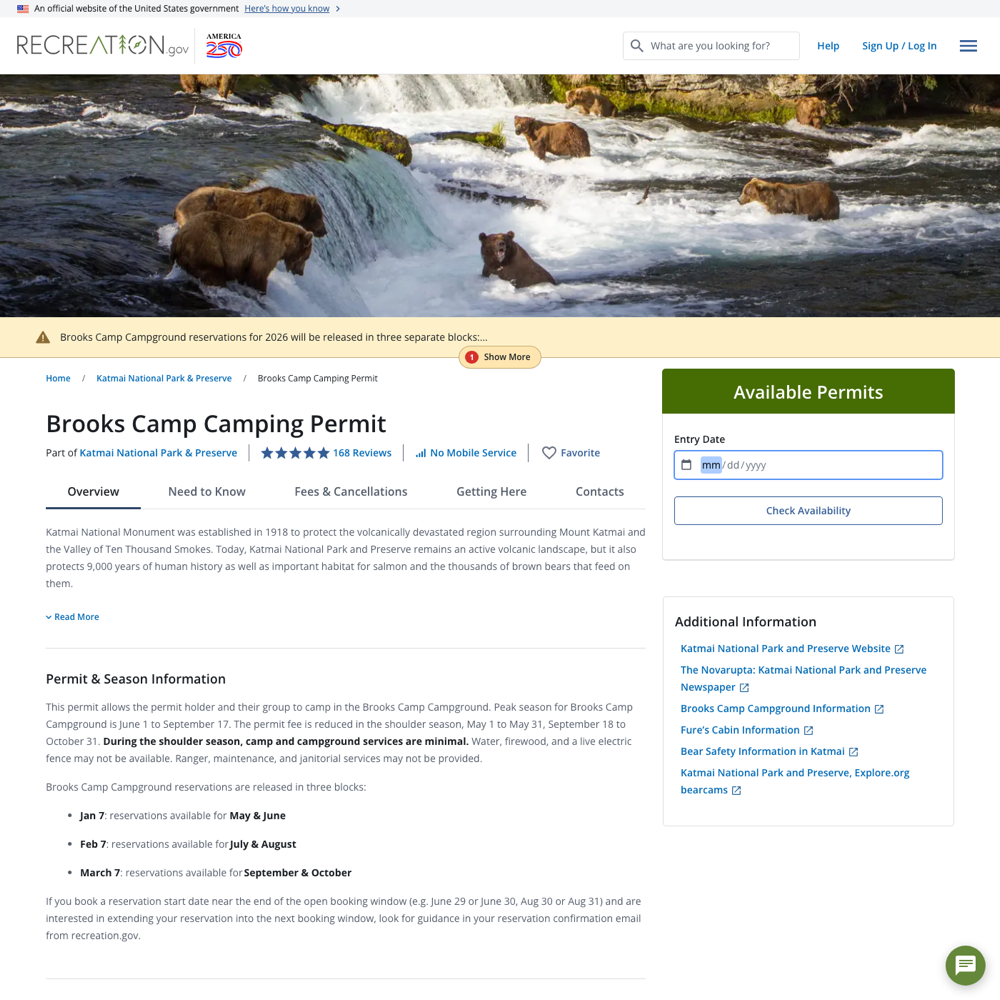}
    \caption{Step 4: \texttt{CLICK} the calendar icon inside the ``Entry Date'' field.}
\end{subfigure}

\vspace{0.2cm}

\begin{subfigure}{0.23\textwidth}
    \includegraphics[width=\textwidth]{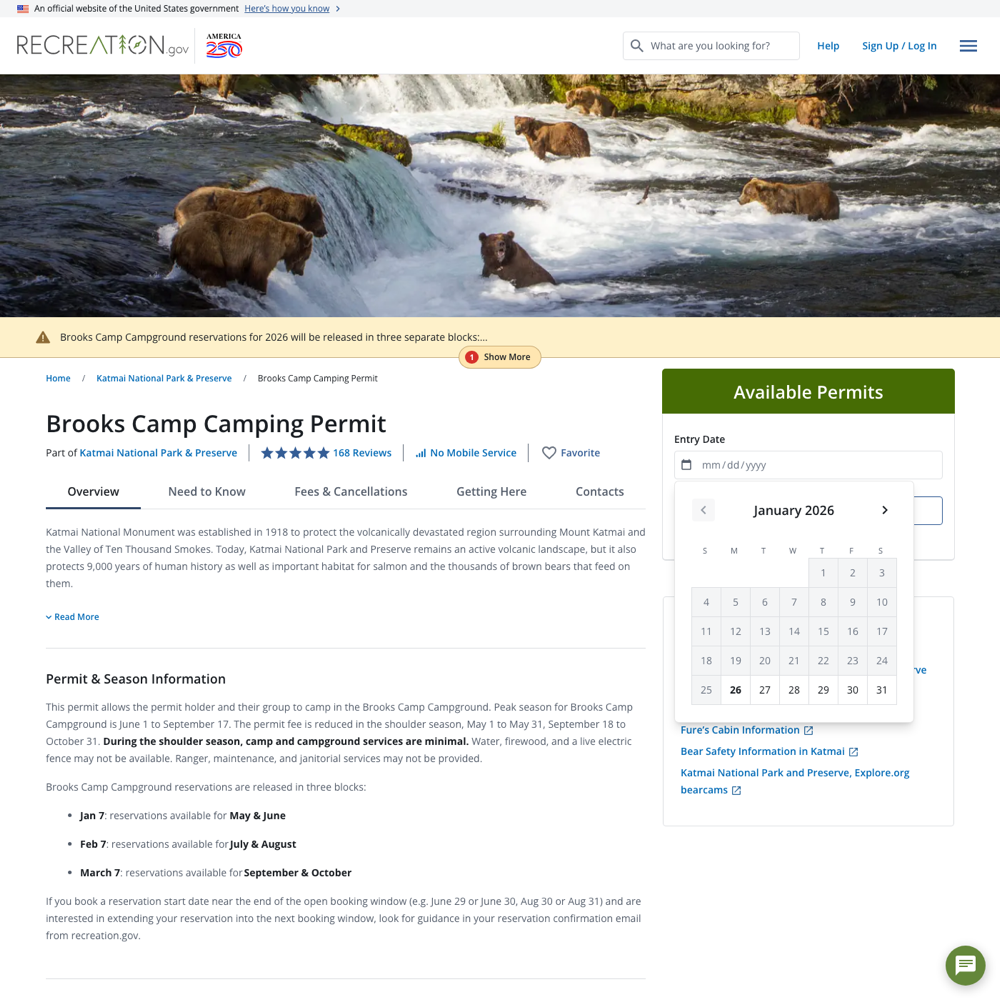}
    \caption{Step 5: \texttt{CLICK} ``31'' to select the target Saturday date.}
\end{subfigure}
\hfill
\begin{subfigure}{0.23\textwidth}
    \includegraphics[width=\textwidth]{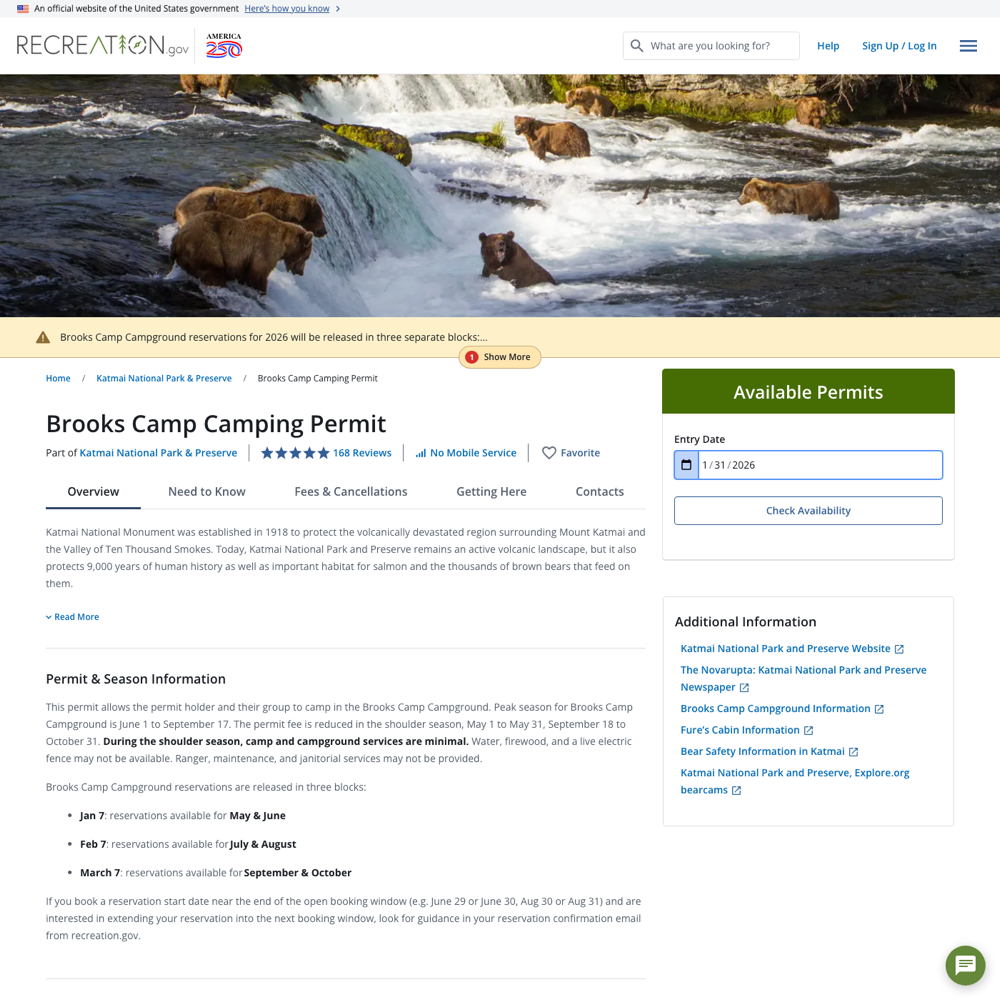}
    \caption{Step 6: \texttt{CLICK} ``Check Availability'' to view permits for the selected date.}
\end{subfigure}
\hfill
\begin{subfigure}{0.23\textwidth}
    \includegraphics[width=\textwidth]{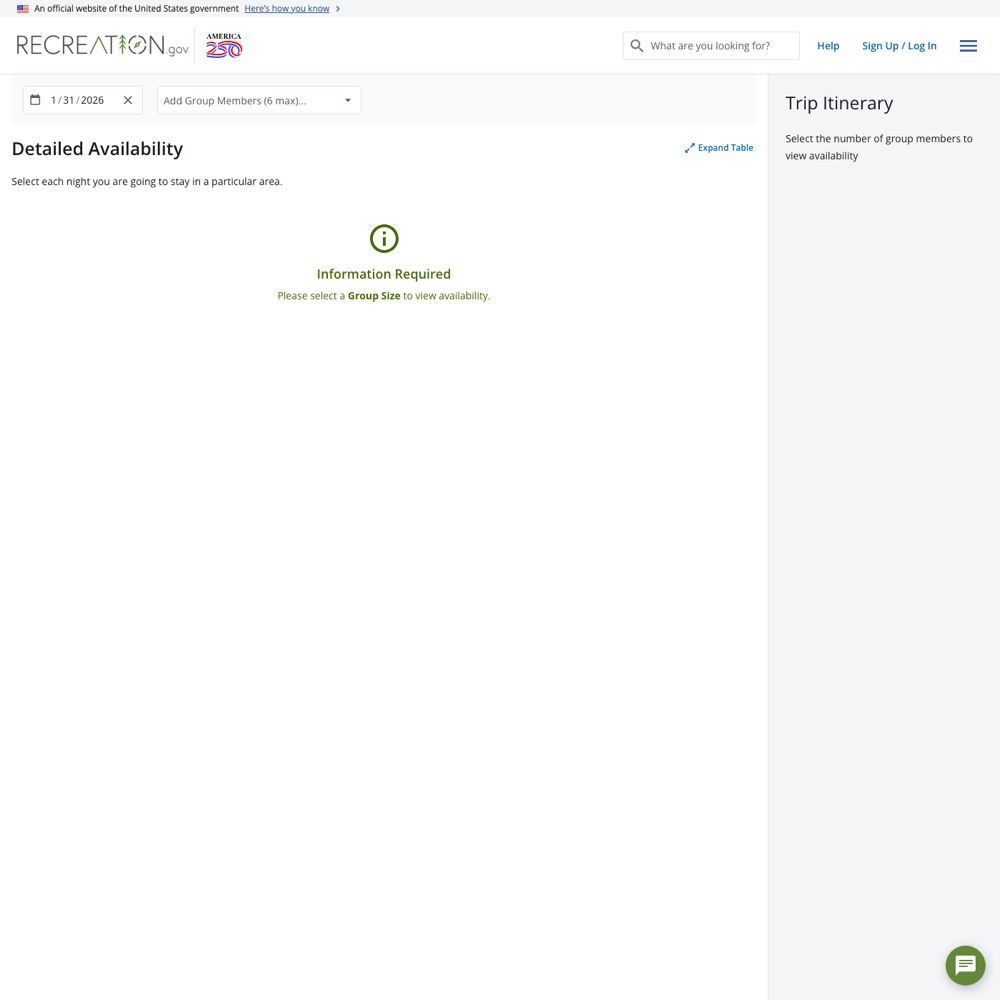}
    \caption{Step 7: \texttt{CLICK} ``Add Group Members'' to set group size.}
\end{subfigure}
\hfill
\begin{subfigure}{0.23\textwidth}
    \includegraphics[width=\textwidth]{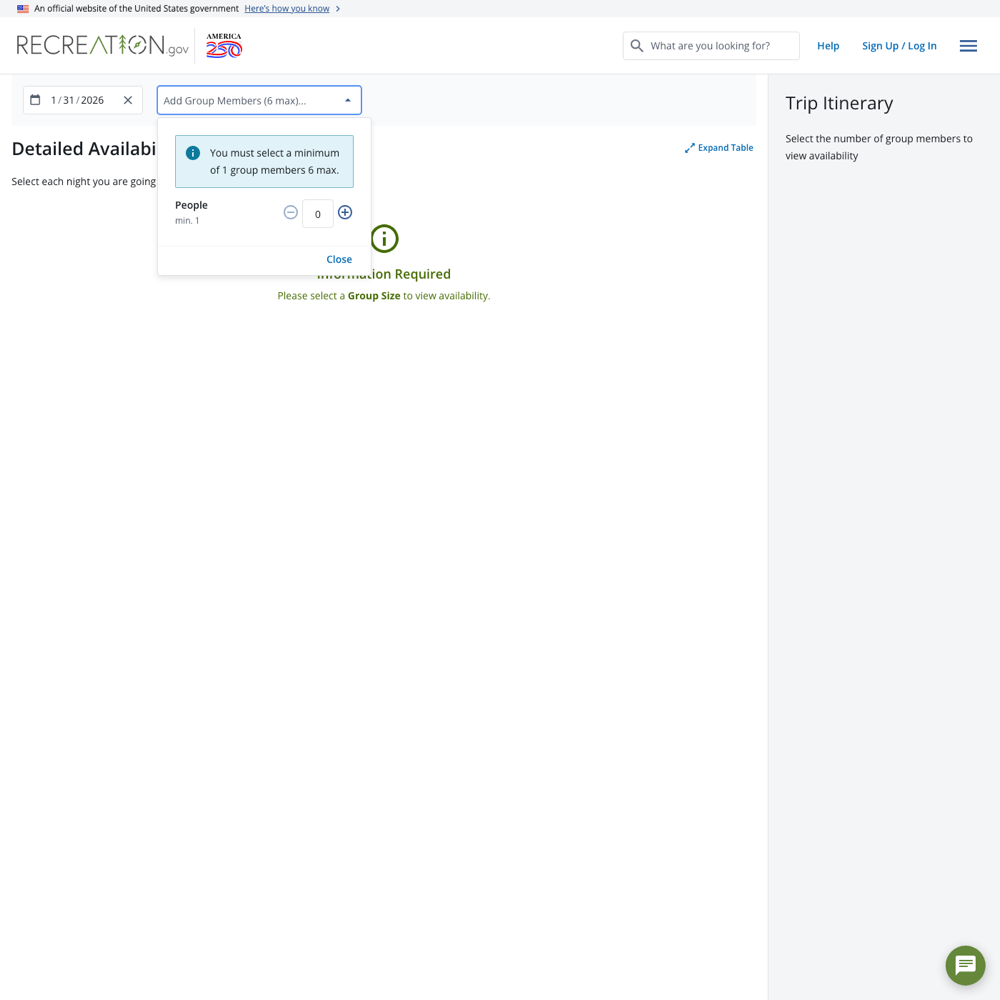}
    \caption{Step 8: \texttt{TYPE} ``4'' into the guest counter input field.}
\end{subfigure}

\vspace{0.2cm}

\begin{subfigure}{0.23\textwidth}
    \includegraphics[width=\textwidth]{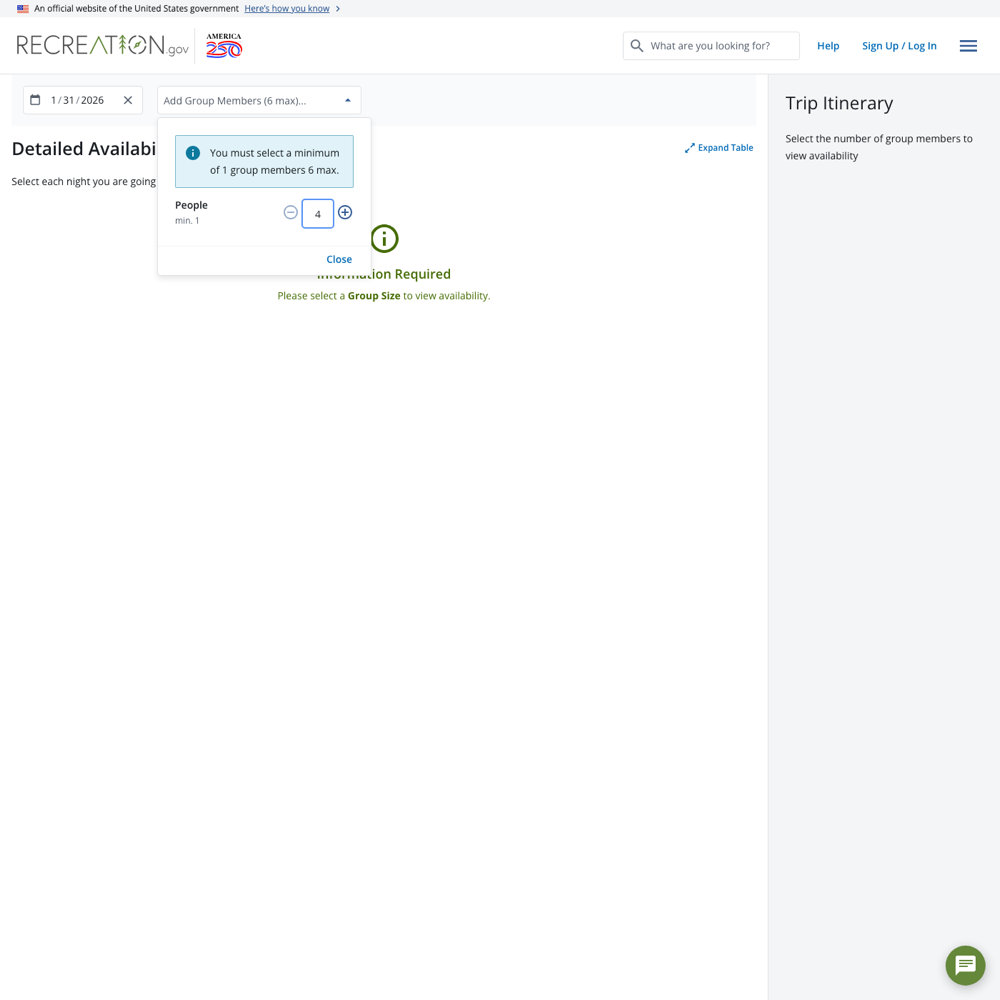}
    \caption{Step 9: \texttt{CLICK} ``Close'' to apply the group size selection.}
\end{subfigure}
\hfill
\begin{subfigure}{0.23\textwidth}
    \includegraphics[width=\textwidth]{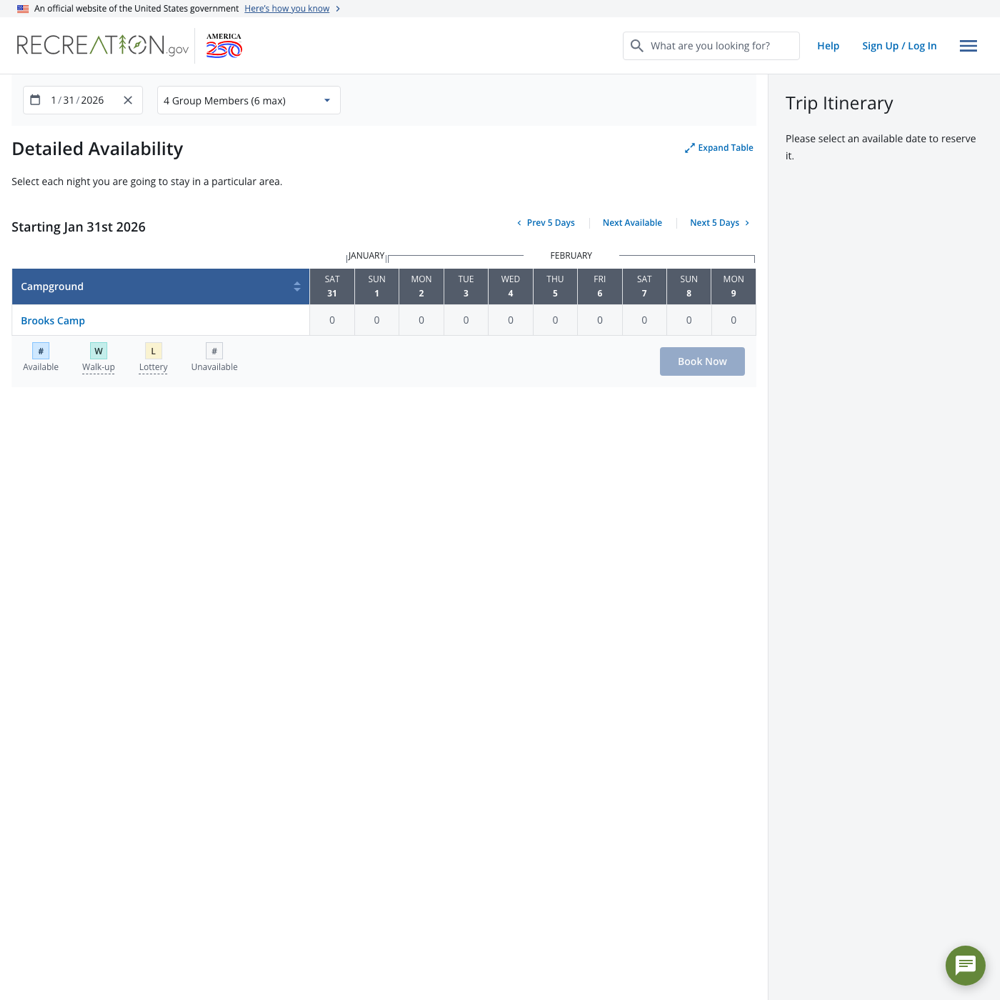}
    \caption{Step 10: \texttt{TERMINATE} after confirming zero availability for the selected date.}
\end{subfigure}
\hfill
\begin{minipage}{0.23\textwidth} \hfill \end{minipage}
\hfill
\begin{minipage}{0.23\textwidth} \hfill \end{minipage}

\vspace{5pt}
\caption{Full 10-step success trajectory of \textsc{Avenir-Web} on \url{recreation.gov}. The agent successfully interacts with dynamic UI components including a calendar picker and a guest counter dropdown. The task concludes with an accurate termination based on the updated availability grid.}
\label{fig:recreation_success}
\end{figure}

As shown in Figure~\ref{fig:recreation_success}, the agent follows a systematic roadmap:
\begin{itemize}
    \item \textbf{Navigation and Search (Steps 1--2):} The agent identifies the search bar, types the target destination, and selects the correct permit link from the auto-suggested results.
    \item \textbf{Dynamic Component Interaction (Steps 3--6):} The agent navigates the complex date-picking interface by first focusing the input field and then using the visual calendar picker to select the specific Saturday. This illustrates the precision of the MoGE module in coordinate-based grounding.
    \item \textbf{Parameter Configuration (Steps 7--9):} The agent opens the guest configuration menu, inputs the numerical group size (``4''), and applies the selection to update the availability grid.
    \item \textbf{Termination (Step 10):} Upon confirming that the availability grid has updated (showing zero spots for the selected parameters), the agent terminates the session with a success status.
\end{itemize}

\paragraph{Complexity and Performance Analysis.} The \url{recreation.gov} trajectory serves as a rigorous stress test for web agents due to the inherent logical complexity of its interface. Unlike static benchmarks, this live task requires navigating a deeply nested UI where information is gated by state-dependent widgets. The difficulty is primarily three-fold: First, the search phase requires disambiguating between various national park services to find the specific ``Brooks Camp Camping Permit,'' a process that often leads to navigational drift in less capable agents. Second, the interaction with the calendar picker represents a significant grounding challenge; the agent must not only locate the grid but also reason over the spatial arrangement of dates to select a specific day (``next Saturday'') while ignoring unavailable or disabled slots. Finally, the task requires maintaining a stable internal state while transitioning from the global search context to the specific parameter configuration menu for group size.

\textsc{Avenir-Web} overcomes these challenges through its dual-layer reasoning architecture. The high-precision visual grounding of the MoGE module allows it to treat the calendar and guest counters as direct visual targets, bypassing the brittle DOM hierarchies that frequently paralyze structural agents. Simultaneously, the synergy between the \textit{Task-Tracking Checklist} and \textit{Adaptive Memory} ensures that the agent remains logically anchored to the user's specific constraints (group of 4, specific date) across multiple state transitions. By completing this high-entropy workflow in a concise 10-step sequence, \textsc{Avenir-Web} demonstrates a level of execution reliability and strategic consistency that bridges the gap between research prototypes and production-ready digital assistants.

\section{Prompts}
\label{sec:prompts}

\subsection{System Prompt}
The system prompt defines the agent's persona, capabilities (tools), and operational rules.

\begin{promptbox}{System Prompt}
\begin{lstlisting}[basicstyle=\ttfamily\scriptsize,breaklines=true]
One action per turn with pixel coordinates.
CLICK: provide 'coordinate' or visible 'text'.
Close/accept blocking modals, overlays, cookie banners first.
Do not repeat actions unless page state visibly changed.
TYPE/SELECT only when target field/dropdown is visible.
KEYBOARD: use 'code' for keys; 'text' for typing; 'CLEAR' to clear active field.
SCROLL: omit coordinates to scroll page; include [x,y] to scroll a container.
If you see potential <select> elements, MUST use 'select' action directly. DO NOT use 'click' to open dropdowns.
When objectives are achieved, TERMINATE with status 'success'.

Strategic guidance:
{strategic_reasoning}

<tools>
{
  "type": "function",
  "function": {
    "name": "browser_use",
    "description": "Single-step browser interaction using pixel coordinates or visible text.",
    "parameters": {
      "type": "object",
      "required": ["action"],
      "properties": {
        "action": {
          "type": "string",
          "enum": ["left_click", "hover", "keyboard", "type", "select", "press_enter", "scroll_up", "scroll_down", "scroll_top", "scroll_bottom", "new_tab", "close_tab", "go_back", "go_forward", "wait", "terminate"]
        },
        "coordinate": {"type": "array", "description": "Normalized [x,y] in 0-1000. REQUIRED for all actions except scroll; omit for scroll only. Include to target a container."},
        "text": {"type": "string", "description": "Visible label or input text. Use 'CLEAR' for keyboard.", "maxLength": 200},
        "code": {"type": "string", "description": "KeyboardEvent.code (e.g., 'PageDown', 'ArrowDown', 'Enter')", "maxLength": 50},
        "clear_first": {"type": "boolean", "description": "Clear active field before typing (type/keyboard)"},
        "press_enter_after": {"type": "boolean", "description": "Press Enter after typing (action=type)"},
        "field": {"type": "string", "description": "Semantic field name (email/search/password/country)", "maxLength": 100},
        "time": {"type": "number", "description": "Seconds to wait"},
        "status": {"type": "string", "enum": ["success", "failure"], "description": "Task status for terminate"},
        "description": {"type": "string", "description": "Short action description (<=200 chars). REQUIRED.", "maxLength": 200}
      }
    }
  }
}
</tools>

Screen: 1000x1000, origin (0,0) top-left.

Rules:
- Do not use GOTO for URL navigation.
- For <select> elements, YOU MUST use 'select' action directly. DO NOT use 'click' to open dropdowns.
- For all actions except scroll actions (scroll_up, scroll_down, scroll_top, scroll_bottom), YOU MUST provide the 'coordinate' parameter with normalized [x,y] values in 0-1000.
- keyboard: use 'code' for keys; 'text' for typing; 'CLEAR' clears the active field.
- **IMPORTANT**: You MUST provide 'coordinate' [x,y] for every CLICK, HOVER, or TYPE action. Do NOT rely on 'text' alone.

Return strictly in <tool_call> tags:
<tool_call>
{"name": "browser_use", "arguments": {"action": "...", ...}}
</tool_call>
\end{lstlisting}
\end{promptbox}

\subsection{User Prompt (Per-Step Input)}
This prompt is constructed at each step to provide the current state and task context.

\begin{promptbox}{User Prompt}
\begin{lstlisting}[basicstyle=\ttfamily\scriptsize,breaklines=true]
Task:
{task}

Pre-step:
Close or accept any cookie/consent banner before other actions.

Strategic guidance:
{strategic_reasoning}

Constraints:
{policy_constraints}

Previous actions:
{previous_actions}

Task progress:
{checklist_context}
\end{lstlisting}
\end{promptbox}

\subsection{Checklist Generation Prompt}
Used to decompose the high-level task into atomic requirements.

\begin{promptbox}{Checklist Generation Prompt}
\begin{lstlisting}[basicstyle=\ttfamily\scriptsize,breaklines=true]
Create 2-6 atomic outcome states based STRICTLY on the task description.

Task: {task_description}

Rules:
1) Each item is an observable goal state (not an action)
2) Max 10 words; short and specific
3) IDs: "requirement_1", "requirement_2", ...
4) Examples: "Size 'blue'", "T-shirt page", "Year: 2022-2023"
5) Status must be lowercase: pending, in_progress, completed, failed
6) DO NOT invent requirements not explicitly mentioned in the task.

Output JSON:
{
    "checklist": [
        {"id": "requirement_1", "description": "First outcome state", "status": "pending"},
        {"id": "requirement_2", "description": "Second outcome state", "status": "pending"}
    ]
}
\end{lstlisting}
\end{promptbox}

\subsection{Checklist Update Prompt}
Used to update the status of checklist items based on the agent's actions.

\begin{promptbox}{Checklist Update Prompt}
\begin{lstlisting}[basicstyle=\ttfamily\scriptsize,breaklines=true]
Update the checklist based on this action:

Action: {action_type} | Success: {success} | Error: {error}

Recent actions:
{history_text}
Page:
{page_state_text}...
Checklist:
{checklist_text}

Update rules:
* completed = fully satisfied
* in_progress = partially done
* pending = not started/reset
* failed = action failed
* Update exactly ONE item per action (most directly affected)
* new_status must be one of: pending, in_progress, completed, failed (lowercase)

Output JSON:
{
    "updates": [
        {"item_id": "requirement_X", "new_status": "pending", "reason": "Brief reason"}
    ]
}
\end{lstlisting}
\end{promptbox}

\subsection{Task Constraints}
Standard safety/policy constraints injected into the User Prompt.

\begin{promptbox}{Task Constraints}
\begin{lstlisting}[basicstyle=\ttfamily\scriptsize,breaklines=true]
Task-specific soft constraints:
- Do NOT attempt to log in, sign in, sign up, or provide credentials.
- If a login/sign-in UI is detected (password fields, 'Sign in', 'Log in', 'Create account'), TERMINATE immediately with status 'failure' and reason 'login prohibited'.
\end{lstlisting}
\end{promptbox}

\section{Experience-Imitation Planning (EIP) Implementation}
\label{sec:appendix_eip}

The \textit{Experience-Imitation Planning (EIP)} module is responsible for generating site-specific roadmaps. Rather than relying on static prompts, the system leverages its integrated search capability to gather real-world knowledge about the target website.

\subsection{Strategic Search and Synthesis}
Before beginning any task, the agent performs a targeted search for the website's help documentation, community forums, or user guides. This allows it to identify specific workflows and interaction patterns that are unique to that site. These findings are then synthesized into a concise, imperative plan that guides subsequent tool calls. This process ensures that the agent is not just guessing, but is following established "best practices" for the specific interface it is interacting with.

\subsection{Execution Flow}
The EIP process follows a strictly defined narrative flow:
\begin{enumerate}
    \item \textbf{Exploration:} The agent starts by searching for the target website's official documentation or relevant community-sourced guidance.
    \item \textbf{Roadmap Generation:} It summarizes the search results into 2--4 actionable sentences, prioritizing visible labels and concrete interaction steps.
    \item \textbf{Strategic Injection:} This high-level roadmap is injected into the main reasoning context, providing a strategic anchor for every subsequent action taken on the page.
\end{enumerate}

\section{Social Impact and Limitations}
\label{sec:social_impact}

\subsection{Ethical Constraints and Anti-Bot Challenges}
In the development and evaluation of \textsc{Avenir-Web}, we maintain a strict ethical stance regarding web interaction and transparency. While proprietary agents such as \textsc{ACT-1} may utilize more aggressive stealth or obfuscation strategies, our framework deliberately avoids the incorporation of CAPTCHA bypass services or header-masking mechanisms. We believe that autonomous agents should operate within the boundaries of a website's intended security protocols. This adherence to transparency is reflected in our results: in our most successful run, approximately 10\% of tasks (31 out of 300) were blocked by the host infrastructure before any actions could be performed, a result that highlights the inherent friction encountered by non-stealthy agents in dynamic web environments.

However, this ethical commitment introduced substantial operational difficulties during our real-world online evaluation:
\begin{itemize}
    \item \textbf{Cloudflare and WAF Challenges:} A significant portion of the websites in the \textsc{Online-Mind2Web} benchmark are protected by advanced Web Application Firewalls (WAFs) like Cloudflare. Without stealth capabilities, our agents were frequently flagged as automated traffic, resulting in persistent "Under Attack" mode challenges or JS-based browser integrity checks that the agent could not resolve autonomously.
    \item \textbf{IP Blocking and Rate Limiting:} Due to the live nature of our evaluation, which often requires repeated visits to the same websites for debugging and trajectory verification, our testing IP addresses were frequently flagged by anti-bot mechanisms (e.g., Cloudflare, Akamai). To ensure consistent evaluation, we utilized a rotating residential proxy service to bypass rate limits.
    \item \textbf{CAPTCHA Barriers:} We encountered numerous instances where legitimate user flows (e.g., account creation or form submission) were gated by CAPTCHAs. Adhering to our policy of not bypassing these mechanisms meant that the agent would terminate the task, negatively impacting our success rate but preserving our commitment to non-evasive research.
\end{itemize}
These challenges highlight a fundamental tension in web agent research: the need for realistic evaluation on the live web versus the necessity of maintaining high ethical and transparency standards. Future work may need to explore "white-listed" evaluation environments or cooperative protocols between agents and website operators.

\subsection{Other Ethical and Safety Considerations}
Beyond anti-bot challenges, the real-world deployment of autonomous web agents raises other significant concerns:
\begin{itemize}
    \item \textbf{Privacy Risks:} Unauthorized access to personal profiles or sensitive user data during automated browsing.
    \item \textbf{Security of Sensitive Operations:} Risks associated with the automation of financial transactions or official form submissions.
    \item \textbf{Harmful Actions:} During our evaluation, we observed that agents could potentially generate harmful actions, necessitating manual safety validation.
\end{itemize}
We release our code strictly for research purposes and firmly oppose any harmful use of this technology.

\subsection{Technical Limitations}
Several technical constraints remain to be addressed in future work:
\begin{itemize}
    \item \textbf{Computational Costs:} Reliance on large-scale MLLMs introduces latency and computational overhead unsuitable for sub-second response applications.
    \item \textbf{Grounding Accuracy:} Hybrid grounding accuracy is bounded by current multimodal backbones (e.g., $\sim$74\% on \textsc{ScreenSpot-Pro}). Even sophisticated strategies like \textsc{Chain-of-Ground} (CoG) utilizing dual-layer \textsc{Gemini 3 Pro} only achieve 75.3\% accuracy, underscoring the difficulty of precise visual interaction.
\end{itemize}

\section{Failure Detection and Recovery Implementation}
\label{sec:appendix_failure_detection}

The detection of operation failure in \textsc{Avenir-Web} is a multi-layered process occurring at four distinct stages, ensuring that tactical errors are caught and propagated for strategic reflection.

\subsection{Trace Analysis of Detection Logic}
The following trace illustrates the internal tool invocations used to determine how action success or failure is detected and recorded within the agent logic.

\begin{promptbox}{Failure Detection Trace}
\begin{lstlisting}[basicstyle=\ttfamily\scriptsize,breaklines=true]
toolName: search_codebase 
query: Find how action success or failure is detected and recorded in agent.py 

toolName: search_codebase 
query: Find where action['success'] or success variable is set in agent.py 

toolName: view_files 
filePath: SeeReAct/src/seeact/agent.py 

toolName: search_by_regex 
query: def.*action 

toolName: search_by_regex 
query: def\s+\w+ 
\end{lstlisting}
\end{promptbox}

\subsection{Multi-Layered Detection Mechanism}

\paragraph{1. Execution-Level Detection}
First, the agent attempts to execute the command in the browser. If the automation fails directly, such as when an element isn't found, a timeout occurs, or a click is intercepted, the error is caught immediately and reported as a failure.

\paragraph{2. State-Change Verification} 
Even if the command runs without error, the agent verifies if it actually changed the page by comparing snapshots before and after the action. It looks for changes in visible text, interactive elements, focus, the URL, scroll position, or modal popups. For actions like clicking or typing, if zero changes are detected, the action is marked as a failure even if no technical error occurred.

\paragraph{3. Action-Specific Checks} 
Certain actions have deeper verification. For example, after typing, the agent reads the value back from the browser to confirm it matches the intended text. If it doesn't match, it is flagged as a failure.

\paragraph{4. Strategic Pattern Analysis} 
Finally, the system analyzes the agent's history of actions to detect potential stalling. If several consecutive identical failures or a high failure rate occur in recent steps, it triggers a warning to reconsider the strategy or terminate the task.